%% file: main.tex
\documentclass[sigconf]{acmart}

\acmSubmissionID{1255}

\usepackage{booktabs} 

\input{IncludesMacros}
\usepackage[ruled]{algorithm2e} 

\SetAlFnt{\small}
\SetAlCapFnt{\small}
\SetAlCapNameFnt{\small}
\SetAlCapHSkip{0pt}

\copyrightyear{2024}
\acmYear{2024}
\setcopyright{acmlicensed}
\acmConference[SIGGRAPH Conference Papers '24]{Special Interest Group on Computer Graphics and Interactive Techniques Conference Conference Papers '24}{July 27-August 1, 2024}{Denver, CO, USA}
\acmBooktitle{Special Interest Group on Computer Graphics and Interactive Techniques Conference Conference Papers '24 (SIGGRAPH Conference Papers '24), July 27-August 1, 2024, Denver, CO, USA}
\acmDOI{10.1145/3641519.3657523}
\acmISBN{979-8-4007-0525-0/24/07}

\begin{document}
\title{Scale-Invariant Monocular Depth Estimation via SSI Depth}

\author{S. Mahdi H. Miangoleh}
\affiliation{
  \institution{Simon Fraser University}
\country{Canada}
}

\author{Mahesh Reddy}
\affiliation{
  \institution{Simon Fraser University}
\country{Canada}
}

\author{Ya\u{g}{\i}z Aksoy}
\affiliation{
  \institution{Simon Fraser University}
\country{Canada}
}

\begin{abstract}
\input{tex/0_abstract}
\input{figures/pipeline}
\end{abstract}



\begin{teaserfigure}
  \centering
   \includegraphics[width=\textwidth]{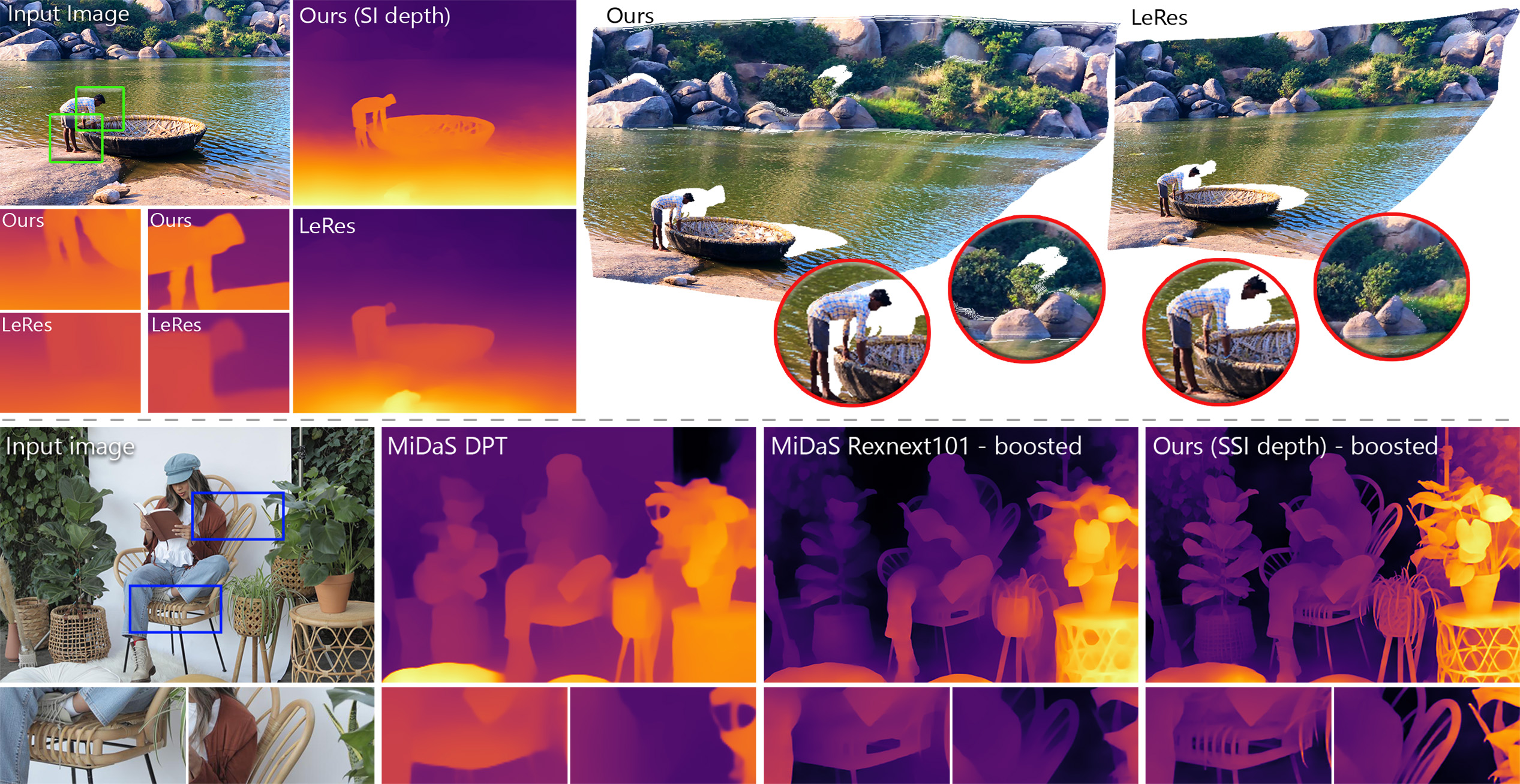}
   \vspace{-0.5cm}
  \caption{(top) We propose a framework to generate high resolution scale-invariant (SI) depth from a single image that can be projected to geometrically accurate point clouds of complex scenes. Our generalization ability comes from formulating SI depth estimation with SSI inputs. (bottom) For this purpose, we introduce a novel scale and shift invariant (SSI) depth estimation formulation that excels in generating intricate details.
  \imagecredits{
  \myhref[darkgray]{https://unsplash.com/photos/man-in-white-shirt-and-blue-denim-jeans-standing-on-brown-wooden-boat-on-body-of-near-near-near-near-Ecd9QETDQwA}{@$\text{Alka Jha}$},  \myhref[darkgray]{https://unsplash.com/photos/woman-in-white-long-sleeve-shirt-and-blue-denim-jeans-sitting-on-brown-wicker-armchair-reading-7b7o3r1DEIg}{@$\text{Joel Muniz}$}
  }
  }
  \label{fig:teaser}
\end{teaserfigure}

\maketitle

\placetextbox{0.14}{0.03}{\includegraphics[width=4cm]{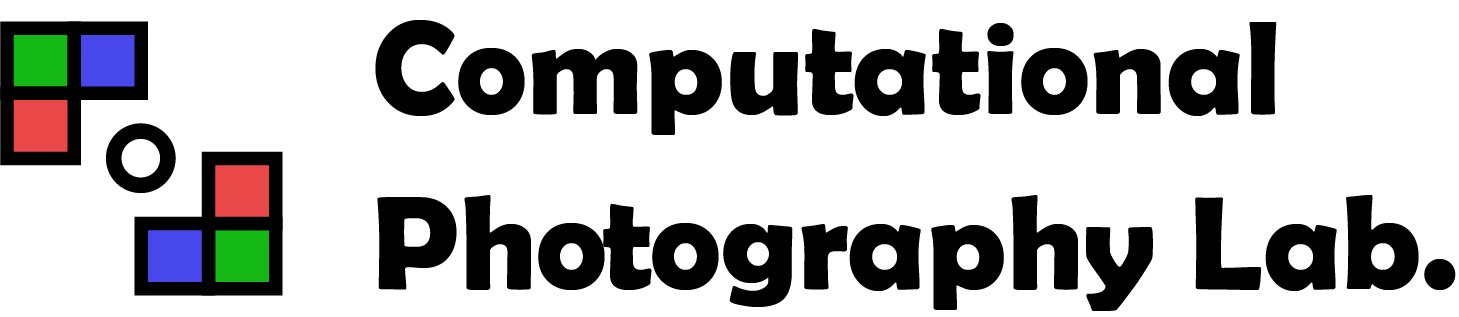}}
\placetextbox{0.85}{0.03}{Find the project web page here:}
\placetextbox{0.85}{0.045}{\textcolor{purple}{\url{https://yaksoy.github.io/sidepth/}}}

\section{Introduction}
\label{sec:intro}

\input{tex/1_intro}

\input{figures/comp_all_depth}

\section{Related Work}
\label{sec:related}
\input{tex/2_related}

\section{High-Resolution SSI Depth Estimation}
\label{sec:ordinal}

\input{tex/3_ordinal}

\section{Scale-Invariant Depth with SSI Inputs}
\label{sec:method}

\input{tex/4_scaleinvariant}

\section{Experiments and Evaluation}
\label{sec:experiments}
\input{tex/5_experiments}

\input{figures/comp_all_depth_SSI_ibims}

\section{Limitations}
\label{sec:limit}
\input{tex/6_limitation}

\section{Conclusion}
\label{sec:conclusion}
\input{tex/7_conclusion}

\begin{acks}
We would like to thank Long Mai and Obumneme Dukor for their help during the early stages of this work, and Chris Careaga for his feedback on the text and the voice-over of our video. We acknowledge the support of the Natural Sciences and Engineering Research Council of Canada (NSERC), [RGPIN-2020-05375].
\end{acks}

\bibliographystyle{ACM-Reference-Format}
\bibliography{references}

\end{document}

%% file: IncludesMacros.tex
\usepackage{atbegshi}
\newcommand{\placetextbox}[3]{
  \setbox0=\hbox{#3}
  \AtBeginShipoutNext{\AtBeginShipoutUpperLeft{%
    \put(\dimexpr#1\paperwidth\relax,-\dimexpr#2\paperheight\relax)
    {\vtop{{\null}\makebox[0pt][c]{#3}}}%
  }}%
}

\newcommand{\uline}[1]{\underline{#1}}

\usepackage{graphics}
\usepackage{array}
\usepackage{multirow}

\usepackage{wrapfig}
\usepackage{hhline}
\usepackage{pifont}
\DeclareMathOperator*{\argmin}{arg\,min} 
\definecolor{myblue}{rgb}{0.3, 0.3, 0.7}

\newcommand{\topscore}[1]{\textcolor{blue}{\textbf{#1}}} 
\newcolumntype{K}[1]{>{\centering\arraybackslash}p{#1}}

\usepackage{duckuments}
\usepackage{tikzducks}
\usepackage{graphicx}
\usepackage{pifont}

\newcommand{\myhref}[3][blue]{\href{#2}{\color{#1}{#3}}}
\usepackage{float}

\newcommand{\ts}{\textsuperscript}

\newcommand{\imagecredits}[1]{\textcolor{darkgray}{\phantom{a} \hfill \small{Image credits: #1}}}

%% file: tex/0_abstract.tex
Existing methods for scale-invariant monocular depth estimation (SI MDE) often struggle due to the complexity of the task, and limited and non-diverse datasets, hindering generalizability in real-world scenarios. This is while shift-and-scale-invariant (SSI) depth estimation, simplifying the task and enabling training with abundant stereo datasets achieves high performance. We present a novel approach that leverages SSI inputs to enhance SI depth estimation, streamlining the network's role and facilitating in-the-wild generalization for SI depth estimation while only using a synthetic dataset for training. Emphasizing the generation of high-resolution details, we introduce a novel sparse ordinal loss that substantially improves detail generation in SSI MDE, addressing critical limitations in existing approaches. Through in-the-wild qualitative examples and zero-shot evaluation we substantiate the practical utility of our approach in computational photography applications, showcasing its ability to generate highly detailed SI depth maps and achieve generalization in diverse scenarios.

%% file: figures/pipeline.tex
\begin{figure*}
    \centering
    \includegraphics[width=\linewidth]{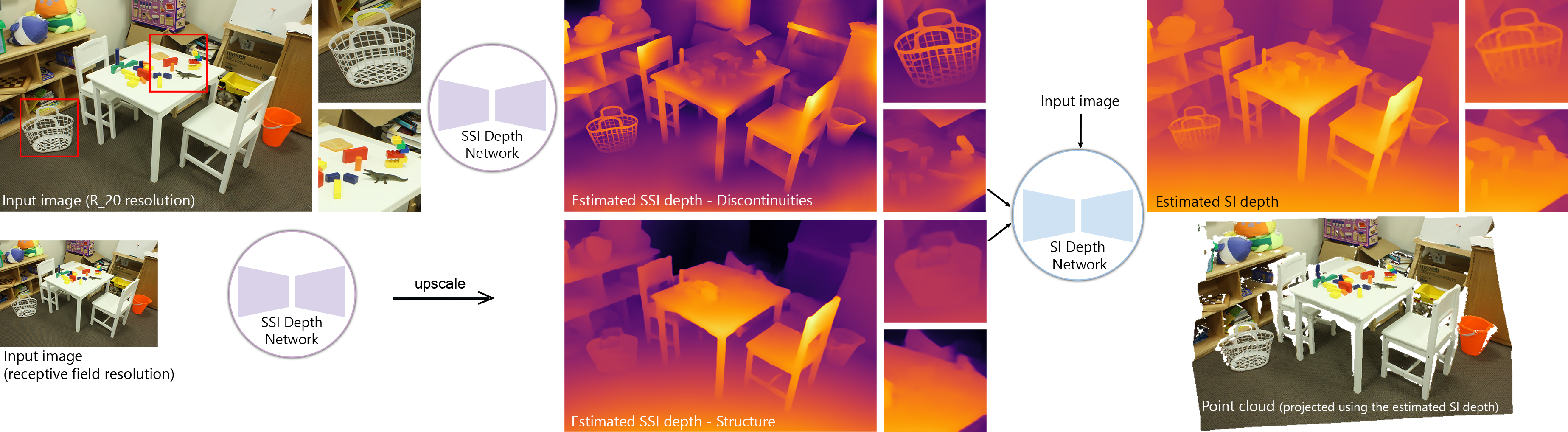}
    \caption{In our framework, we employ a low-resolution SSI depth estimation to capture the rough scene structure, and a high-resolution SSI depth estimation representing sharp depth discontinuities.
    Feeding this rich structural information to the SI network, we regress the high-resolution scale-invariant monocular depth that can be projected into geometrically accurate point clouds.
    \imagecredits{
    Middlebury dataset~\cite{scharstein2014high}
    }
    }
    
    \label{fig:pipeline}
\end{figure*}

%% file: tex/1_intro.tex
Monocular depth estimation (MDE) is a fundamental mid-level computer vision problem and a critical part of computational photography pipelines such as 3D photography, free view-point rendering, and depth-based editing on individual photographs \cite{niklaus20193d,shih20203d,Peng2022BokehMe,wadhwa2018synthetic}. 
Lacking geometric cues available in multi-view reconstruction formulations, MDE is a challenging high-level problem that requires reasoning about monocular depth cues such as occlusions, relative object size, and converging lines. 
The challenge is further enhanced for computer graphics applications with the requirement of high-resolution estimations and in-the-wild generalization.

MDE can be defined as the estimation of the physical distance of every pixel to the camera, which is referred to as \emph{metric depth} that requires the focal length of the camera as well as semantic knowledge of the size of objects. 
The scene geometry, however, can be captured up to a scale with an unknown focal length reflecting the inherent scale invariance in image formation. 
The estimation of this geometric depth is referred to as \emph{scale-invariant (SI) MDE}. 
While the metric scale is required for robotics applications such as autonomous driving, computational photography applications only require the geometric SI depth for rendering. 
In this work, we focus on achieving high-resolution SI MDE in in-the-wild and complex scenes.

Due to the lack of high-resolution, large-scale, and diverse training datasets for SI depth, earlier methods have failed to achieve the boundary accuracy and generalizability demanded by photography applications. 
To address the generalizability challenge, several works~\cite{yin2019enforcing,leres,midas} define the MDE problem in the disparity space coming from stereo pairs with unknown baselines. 
This stereo MDE is referred to as \emph{scale-and-shift-invariant (SSI) depth}, reflecting the arbitrary shift from the true geometry inherent in stereo pair disparities. 
With the abundance of stereo training datasets, SSI depth is shown to have better generalization compared to SI MDE. 
SSI MDE also generates better details at high resolutions \cite{bmd}, but is insufficient for computer graphics applications due to the loss of geometric accuracy.

In this work, we propose an SI MDE pipeline, visualized in Figure~\ref{fig:pipeline}, that makes use of abundant stereo datasets for in-the-wild high-resolution geometric depth estimation. 
Our pipeline consists of an initial SSI depth estimation, the results of which are fed to a second SI depth estimation network. 
We first develop a novel sparse loss to improve SSI MDE performance in detail generation and boundary accuracy. 
We show that our SSI depth estimation outperforms the current state-of-the-art, allowing highly detailed estimation of depth discontinuities even in complex scenes as Figure~\ref{fig:teaser} demonstrates. 
We use our SSI MDE network to generate an overall scene structure and high-resolution depth discontinuities to be given to our SI network as input.

Given rich structural information in the form of SSI depth, the task of our SI MDE network gets simplified into the enforcement of geometric constraints. 
This simplified task definition narrows the domain gap between synthetic datasets and in-the-wild images. 
We show that with the generalizable SSI depth used as input, in-the-wild geometric depth estimation can be achieved using only synthetic SI depth datasets for training. 
Effectively, our two-step pipeline allows us to harness the advantages of SSI MDE to generate highly detailed geometry from a single image in a wide variety of scenes as shown in Figure~\ref{fig:comparison:all:depth}. 
We demonstrate the practical use of our methodology through qualitative examples and 3D computational photography applications in the supplementary material.

%% file: figures/comp_all_depth.tex
\begin{figure*}
    \centering
    \includegraphics[width=\linewidth]{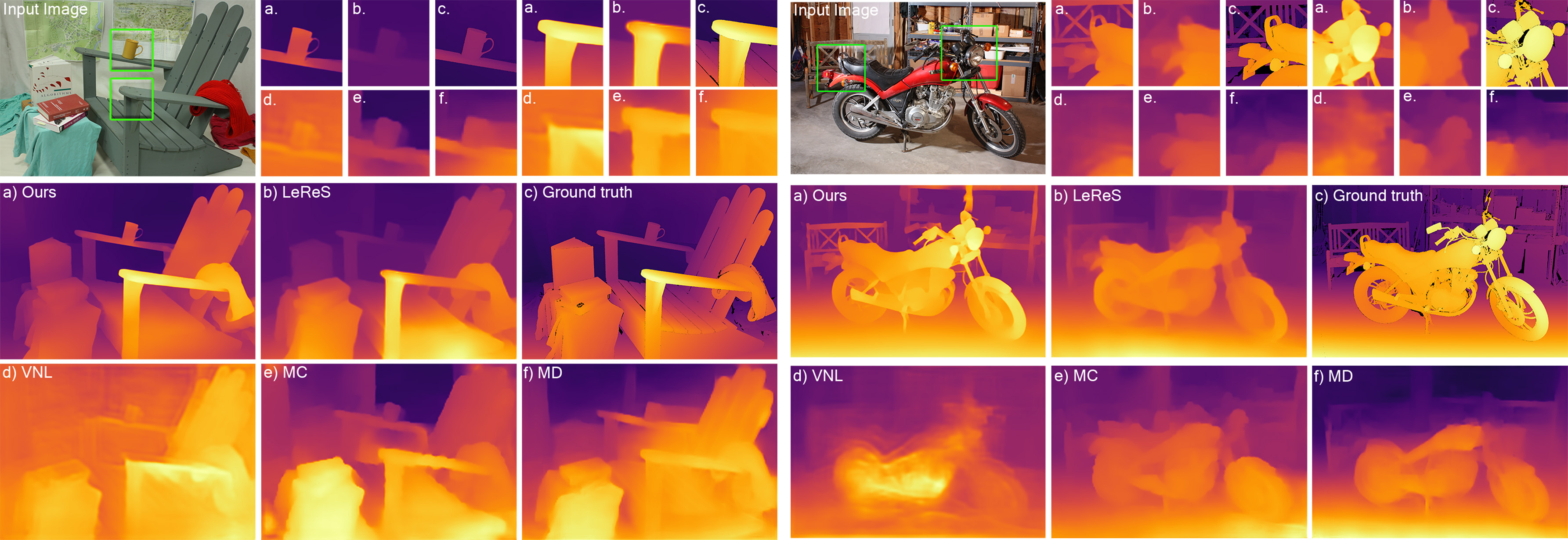}
    \caption{Qualitative comparison of scale-invariant networks on the Middlebury dataset~\cite{scharstein2014high}. Our scale-invariant network exhibits superior performance in capturing intricate objects with higher levels of depth details compared to the state-of-the-art. 
    }
    \label{fig:comparison:all:depth}
\end{figure*}

%% file: tex/2_related.tex
Monocular depth estimation (MDE) is a high-level task that requires reasoning about monocular depth cues such as occlusions, perspective, and relative size of objects. 
Hence, modern MDE approaches are overwhelmingly data-driven to implicitly learn the depth cues \cite{eigen2014depth, godard2017unsupervised, Ramamonjisoa_2020_CVPR, wang2020cliffnet, wong2019bilateral, zheng2018t2net}. 

\subsubsection*{Scale-and-shift-invariant MDE}
In-the-wild MDE requires large training datasets for better generalization to in-the-wild images. 
However, due to the difficulty in capturing metric depth ground-truth at high-resolution at scale, there only exist a few real-world datasets for this task. 
In order to use datasets with stereo image pairs to extend the available training data, \citet{midas} develop a scale-and-shift invariant (SSI) depth formulation and train their network in the disparity space.
The relaxed formulation and the extended set of image-ground truth pairs with novel loss functions improve the accuracy of the estimated SSI depth and allow generalization to in-the-wild images, but the unknown shift still needs to be recovered for geometric reconstruction. 
\citet{bmd} propose a boosting framework and demonstrate that high-resolution SSI depth with rich details can be achieved using CNN-based SSI models through inference at two different resolutions. 
\citet{yin2019enforcing,yin2021virtual} define a virtual plane and define a loss on its surface normal and combine it with an SSI loss in the depth space. \citet{leres} resolves the SSI ambiguity by normalizing the depth distribution per image with a Gaussian assumption. 
\citet{dpt} propose a novel transformer architecture for dense depth estimation that benefits from the higher learning ability of transformer architectures. 
\citet{depthanything} exploit unlabeled data and pretrained depth models to generate pseudo ground truth to train a student model that is more effective than the original teacher model.
We introduce a novel sparse ordinal loss and demonstrate that combining our sparse loss with the dense SSI loss enhances detail generation for our SSI model.

\subsubsection*{Ordinal MDE}
As \citet{midlevel} explore, estimating whether a pixel is closer to the camera than the other without enforcing geometric constraints leads to better performance in estimating depth discontinuities. 
Several works \cite{chen2016single,chen2019learning,sgr,xian2018monocular} aim for dense estimation of ordinal depth using sparse ranking loss functions that only enforce the correct ordering of pixels. \citet{sgr} shows that they generate MDE with more high-resolution details when compared to MiDaS \cite{midas} which still encodes the geometry. 
As discussed in Section~\ref{sec:ordinal}, traditional sparse ranking loss cannot be combined with SSI loss. However, our sparse loss is compatible with SSI loss, thereby enabling the detail-generating capability of the sparse loss for SSI depth estimation.

\subsection{Scale-invariant MDE}
The scale-invariant (SI) depth estimation networks require geometrically consistent depth maps that are a scale away from the true depth. 
The earlier data-driven methods approached MDE geometrically through scale-invariant loss definition~\cite{eigen2014depth,eigen2015predicting,li2018megadepth}. 
Also, the geometrically consistent scale-invariant depth can be used to reliably estimate the surface normals as well. 
This connection has also been exploited to train MDE for scale-invariant estimation using surface normals~\cite{chen2017surface}. 
Due to the complex nature of SI depth estimation and the constrained capacity of neural networks, models attempting to directly estimate SI depth struggle to generate fine details. Additionally, these models often face limitations due to dataset constraints, being typically trained on a single dataset, which reduces their ability to generalize to in-the-wild images.

\citet{leres} proposes to estimate SI depth by estimating the unknown shift in SSI depth using a network trained on point clouds. This allows benefiting from the SSI depth to achieve better generalization for SI depth estimation. 
However, as shown in Figure~\ref{fig:teaser}, their method fails to generate detailed geometry for complex scenes. This deficiency stems from its reliance on the geometry estimated from a single pass through an SSI network. 

To achieve a high level of details and leverage the generalizability of SSI for SI MDE, we utilize an SSI depth network with a CNN backbone and propose feeding SSI depth at both low and high resolutions as input to a dense SI depth estimation model. By incorporating the local details generated from high-resolution SSI depth, in addition to the structure of the low-resolution SSI depth, our method demonstrates the capability to generate highly detailed SI depth. We utilize a publicly available synthetic dataset to train our SI model, demonstrating that, owing to the simplified task of SI depth estimation by feeding SSI depth to it, our model can generalize effectively to diverse scenes despite being trained on a single dataset.

\subsubsection*{Metric MDE}

Metric MDE directly regresses metric depth. To enhance this process, significant efforts have been made to refine network architectures \cite{Chen2019structure-aware, eigen2014depth}, incorporate CRFs \cite{yuan2022newcrfs}, or reformulate continuous depth regression as a classification task \cite{bhat2021adabins, bhat2022localbin, dorn2018}. 
To simplify the task, \citet{jun2022depth, Jae2019depthusingrel} decompose metric depth into ordinal features and aim to estimate the full metric depth using these features. \citet{zoedepth} utilize low-resolution SSI depth to estimate metric depth. 
\citet{patchfusion} employed a brute-force, patch-based approach to estimate depth, succeeding in generating highly detailed depth maps. 
However, this iterative approach results in slower runtime compared to the state of the art. Ultimately, when in-the-wild images taken with varying focal lengths cause the depth regressed by these methods to differ from the actual metric depth by a scale factor. 
This discrepancy reduces their effectiveness to that of a scale-invariant (SI) depth estimation model, despite significant network capacity being dedicated to training for metric depth estimation.

%% file: tex/3_ordinal.tex
Seeking high-resolution scale-invariant monocular depth estimation (SI MDE), our approach unfolds in two key steps. The initial stage involves extracting the overall structure and high-resolution depth discontinuities through the application of the scale-and-shift invariant (SSI) formulation. This extracted information serves as the input to our subsequent scale-invariant depth estimation network.

The concept of SSI monocular depth estimation was originally introduced by \citet{midas} as a more flexible alternative to traditional SI MDE. It allows for training SSI networks on stereo datasets with unknown baselines while also utilizing standard SI ground-truth. \citet{bmd} further demonstrated the feasibility of achieving high-resolution SSI estimations by combining multi-resolution outputs. Exploiting these dual advantages --generalizability and accurate depth discontinuities-- our focus lies in enhancing the high-resolution accuracy of the SSI formulation.

To achieve this goal, we introduce a novel sparse ordinal loss for SSI training. This loss contributes to the improved high-resolution performance of our SSI formulation, a critical component in generating detailed SI depth estimations in the subsequent stages of our pipeline.

\subsection{Sparse ordinal loss}

SSI depth estimation is characterized by its scale and shift-invariant loss \cite{midas} defined in the disparity space:
\begin{equation}
    \mathcal{L}_{ssi} = \frac{1}{N} \sum_{i}^{N} (f(O_i) - D^*_i)^2,
\end{equation}
where $O$ is the estimated disparity, $D^*$ is the ground-truth, and
\begin{equation}
    f(x) = ax + b \quad 
    (a, b) = \argmin_{a, b} \sum_i (f(O_i) - D^*_i)^2, \quad a>0
\end{equation}
is a linear function parameters of which are estimated for each individual estimation during training. 
This formulation is particularly useful as it allows the use of both geometric ground-truth as well as disparities estimated from stereo pairs with unknown baseline distance for training.

The SSI loss is a global function, instilling coherence in the depth estimation structure. 
However, the sole use of the SSI loss does not allow the network to generate sharp depth discontinuities when compared to sparse ordinal formulations.
To enhance the emphasis on sharp depth discontinuities, we introduce a sparse ordinal loss, working in tandem with the dense SSI loss, to enforce the correct ordering of pixel pairs in the depth space.
For a given pixel pair $(i,j)$, we define our ordinal loss as:
\begin{equation} 
    \mathcal{L}_o(i,j) = 
    \begin{cases}
        (\Delta O_{ij})^2 \hfill \text{if }  |\Delta \hat{O}_{ij}| < \delta \\
        \texttt{ReLU}\left(-\Delta O_{ij} \times \texttt{sgn}(\Delta \hat{O}_{ij})\right) \text{ otherwise}
    \end{cases}
\end{equation}
Here, $O$ and $\hat{O}$ represent the estimated and ground-truth disparity, respectively, and $\Delta O_{ij} = O_i - O_j$. The term $\delta = 0.01$ is a small threshold, defining when two points are considered to be at the same depth. For pixels at different depths, we apply a linear loss only when the estimated ordering of the pair diverges from the ground-truth. Conversely, for pixels at similar ground-truth depths, we apply an $L_2$ loss, encouraging estimations to be similar.

\input{figures/rankinglossplot}

As discussed in Section~\ref{sec:experiments:ablation:ssi}, our sparse loss significantly enhances the edge accuracy of SSI estimations. This improvement aligns with the advantages of other sparse losses observed in the realm of relative depth literature, particularly the ranking loss by \citet{chen2016single} and its subsequent use by others \cite{chen2019learning,sgr,xian2018monocular}. 
A drawback of the sparse ranking loss, as illustrated in Figure~\ref{fig:method:rankingLossPlot}, is their non-zero contribution even when the pixel ordering is correct. This characteristic leads to a conflict when combined with the SSI loss, rendering their joint use impractical. In contrast, our ordinal loss is carefully defined to circumvent such conflicts and enabling seamless integration with the SSI loss.

Following \citet{chen2016single}, we compute our sparse ordinal loss over 2500 randomly sampled pixel pairs over the image, $\mathcal{L}_{so} = \sum_{\forall (i,j)} \mathcal{L}_o(i,j)$. 
We define our final loss with the SSI and sparse ordinal losses, as well as the multi-scale gradient loss $\mathcal{L}_{ssig}$ \cite{li2018megadepth} as an edge-aware smoothness metric:
\begin{equation}
    \mathcal{L}_{ssiNet} = \lambda_{ssi} \mathcal{L}_{ssi} + \lambda_{so} \mathcal{L}_{so} + \lambda_{ssig} \mathcal{L}_{ssig},
\end{equation}
where $\lambda_{ssi}=3$, $\lambda_{so}=1$, and $\lambda_{ssig}=0.1$.

\subsection{Training details}

We follow \citet{midas} and adapt the network architecture from \citet{xian2018monocular} with a ResNeXt101 \cite{xie2017aggregated} feature extractor with weakly supervised learning weights \cite{mahajan2018exploring} as initialization. We use the sigmoid activation to predict the ordinal inverse-depth, i.e. disparity, in $[0, 1]$ and train the network with Adam optimizer for 30000 iterations with a learning rate of $10^{-3}$. More details are provided in the supplementary material.

We train our ordinal network on a diverse set of datasets for better generalization.  
From the Omnidata framework \cite{eftekhar2021omnidata}, we use the Hypersim \cite{roberts2021hypersim}, Replica~\cite{straub2019replica} and Replica+GSO~\cite{choi2016large} datasets. 
We also use the synthetic OpenRooms \cite{li2021openrooms}, TartanAir \cite{wang2020tartanair}, and FSVG \cite{krahenbuhl2018free} datasets. 
In addition, we use the real-world stereo datasets HRWSI~\cite{sgr} and Holopix50k~\cite{hua2020holopix50k} where we compute the disparity maps using RAFT~\cite{teed2020raft} and use the sky segments from Mask2Former~\cite{cheng2021masked}.
This wide range of real-world datasets ensures that the SSI inputs we generate for our SI network are reliable in a wide range of real-world scenarios.

%% file: figures/rankinglossplot.tex
\begin{figure}[!t]
    \centering
    \includegraphics[width=0.8\linewidth]{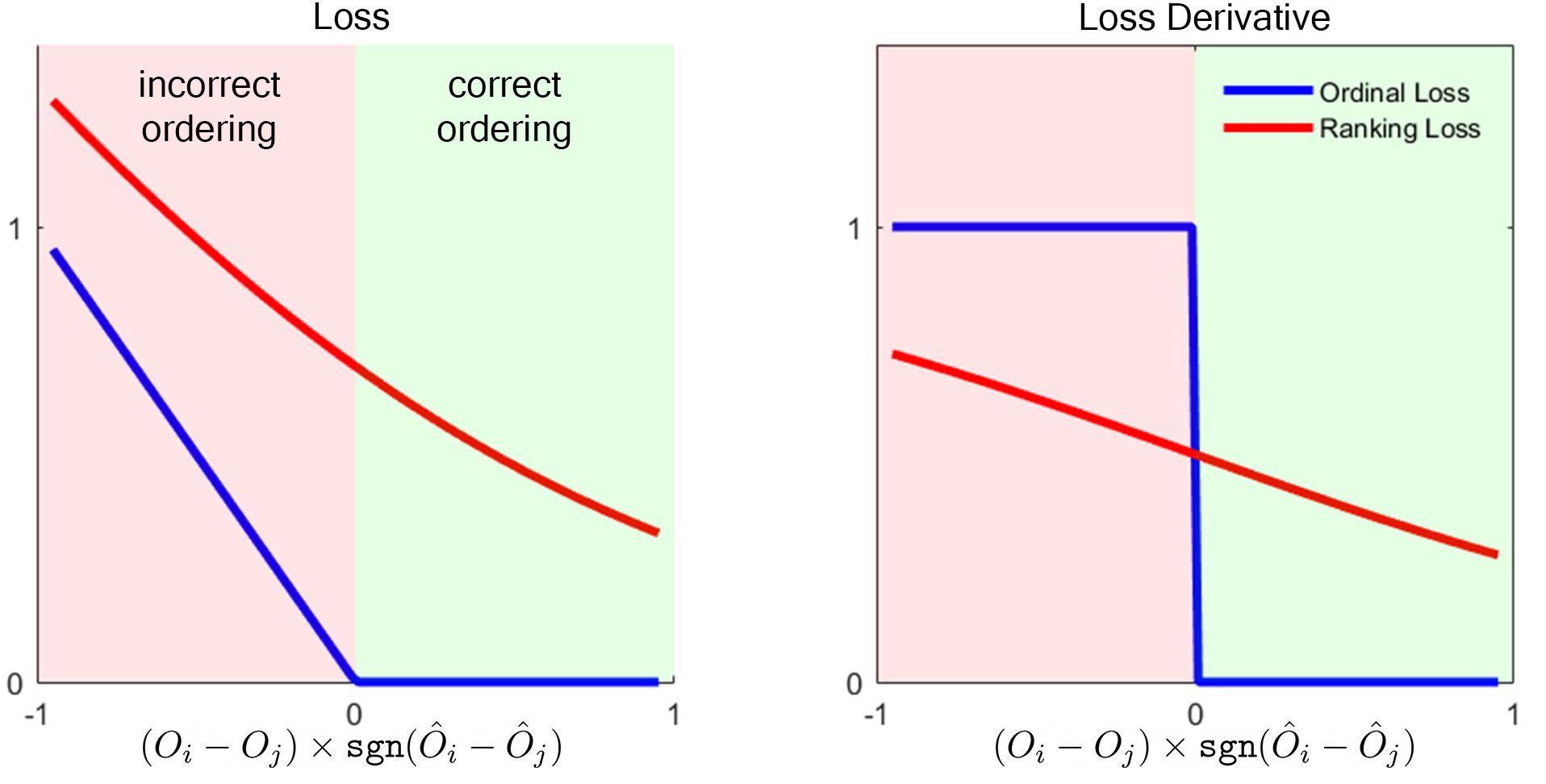}
    \vspace{-0.25cm}
    \caption{
    The plot of our ordinal loss and the ranking loss~\cite{chen2016single}.
    The ranking loss assigns a high penalty for correctly ordered pairs, while we only apply a penalty for incorrectly ordered pairs.
    }
    \vspace{-0.25cm}
    \label{fig:method:rankingLossPlot}
\end{figure}

%% file: tex/4_scaleinvariant.tex
We redefine the scale-invariant monocular depth estimation (SI MDE) problem by incorporating scale-and-shift invariant (SSI) inputs. The network receives two SSI inputs concatenated with the input image, forming an input of dimensions $h \times w \times 5$, where $h$ and $w$ represent the height and width of the input image, respectively. The first SSI input, denoted as $\mathcal{O}^L$, is computed at the receptive field size of the SSI network, offering an overall depiction of the scene's structure. The second SSI input, $\mathcal{O}^H$, is generated at a higher resolution, capturing intricate depth discontinuities. 
The resolution of $\mathcal{O}^H$ estimation is selected based on the image content using the $\mathcal{R}_{20}$ measurement by \citet{bmd} using the local edge density.
This formulation streamlines the task for the SI MDE network, focusing on enforcing geometric constraints using the provided overall structure and high-resolution depth information.

Benefiting from a diverse training dataset for our SSI network, the SSI inputs exhibit robust generalization to in-the-wild imagery. By simplifying the task definition for the SI network and leveraging generalizable SSI inputs, we demonstrate the feasibility of achieving in-the-wild high-resolution SI depth estimation via synthetic-only training.

\input{tables/si_zeroshot}

\subsection{Scale ambiguity}

The inherent scale ambiguity in SI depth formulations necessitates reliance on scale-invariant losses during network training. These losses, enforcing a least squares fit between network estimations and ground truth, face challenges in determining scale during the initial training phases due to inherent inaccuracies in under-trained networks.

In our framework with scale-and-shift invariant (SSI) inputs, the globally consistent low-resolution SSI estimation, denoted as $\mathcal{O}^L$, acts as a stable reference for our SI MDE network. Sequential training, starting with the SSI network, ensures a stable least-squares fit between $\mathcal{O}^L$ and ground truth. Utilizing this low-resolution input establishes the ground truth's arbitrary scale with stability during early training, addressing challenges posed by scale ambiguity.

In accordance with the SSI estimation, we formulate SI depth estimation in the inverse depth space. To maintain stability, we fix the ground truth's arbitrary scale throughout training using:
\begin{equation}
    c = \argmin_s \sum_i \left(s \hat{D}^*_i - \mathcal{O}^L_i \right)^2, \quad \hat{D} = c \hat{D}^*,
\end{equation}
Here, $\hat{D}^*$ and $\hat{D}$ represent the original and scale-adjusted ground truth inverse depth. Simultaneously, we align the average scale of the high-resolution SSI input $\mathcal{O}^H$ with that of $\mathcal{O}^L$, ensuring consistent scales for input and output variables. Fixing the arbitrary scale provides a foundation for defining dense losses without requiring scale invariance, enhancing stability and effectiveness in training.

\subsection{Loss functions}

Setting the arbitrary scale using $\mathcal{O}^L$, we employ a straightforward $L_1$ loss ($\mathcal{L}_{d}$) on estimated depth values and scale-adjusted ground truth as well as a multi-scale gradient loss~\cite{li2018megadepth} ($\mathcal{L}_{dg}$) for spatial coherency.

Additionally, following \citet{chen2017surface}, we include a surface normal loss ($\mathcal{L}_n$), defined by the cosine-similarity between normals computed from estimated depth ($n$) and ground-truth normals ($\hat{n}$).
%
We also incorporate a multi-scale gradient loss on surface normals, defined effectively on the second derivative of SI depth, promoting better curvature in estimations. This enhances the geometric representation and spatially coherent surface normals:
\begin{equation}
    \mathcal{L}_{ng} = \frac{1}{NM} \sum_m \sum_i \left( \nabla\hat{n}^m_i - \nabla{n}^m_i \right)^2.
\end{equation}

where $\nabla{n}^m$ is dimensions-wise gradients of the surface normal and $M$ is the number of scales. Our overall loss, combines each component with appropriate weights:
\begin{equation}
    \mathcal{L}_{siNet} = \lambda_d \mathcal{L}_d + \lambda_{dg} \mathcal{L}_{dg}
                            + \lambda_n \mathcal{L}_n + \lambda_{ng} \mathcal{L}_{ng},
\end{equation}
where $\lambda_{d}=1$, $\lambda_{dg}=0.5$, $\lambda_{n}=0.1$, and $\lambda_{ng}=0.01$.

\subsection{Training details}

We adopt the architecture from~\citet{xian2018monocular} with EfficientNet-b7~\cite{tan2019efficientnet} as the backbone for our scale-invariant depth estimation network. Training resolution is $1024 \times 1024$, and during inference, we resize to a maximum dimension of 1024 pixels while maintaining aspect ratio. We use the ADAM optimizer (learning rate of $1e-3$) for 30 epochs with a batch size of 2 to train the network.

Given the scarcity of high-resolution real-world datasets with scale-invariant (SI) depth ground truth, we exclusively train on the synthetic dataset Hypersim~\cite{roberts2021hypersim}. This dataset offers high-resolution ground truth for both SI depth and surface normals. To avoid over-fitting to the intrinsic parameters of this dataset we use diverse crop augmentations as described in supplementary material. The simplified task for the SI network and generalization from SSI inputs enable in-the-wild SI monocular depth estimation by training solely on this synthetic indoor dataset, as demonstrated in qualitative evaluations.

%% file: tables/si_zeroshot.tex
\begin{table*}[!t]
\caption{Quantitative evaluation for scale-invariant depth estimation. We are reporting surface normal metric with $t=11.25^{\circ}$. }
\label{tab:si_zeroshot}
\vspace{-0.25cm}
\resizebox{\textwidth}{!}{%
{\renewcommand{\arraystretch}{1.2}
    \begin{tabular}{l|cccccc|cccccccc|cccccc}
    \hline
    Methods 
    & \multicolumn{6}{c|}{Middlebury} 
    & \multicolumn{8}{c|}{iBims-1} 
    & \multicolumn{6}{c}{DIODE} 
    \\

    \multicolumn{1}{c|}{} 
    & \multicolumn{3}{c}{Structure and Shape} 
    & \multicolumn{2}{c}{Surface Normal} 
    & \multicolumn{1}{c|}{Edges} 

    & \multicolumn{3}{c}{Structure and Shape} 
    & \multicolumn{2}{c}{Surface Normal} 
    & \multicolumn{3}{c|}{Edges} 

    & \multicolumn{3}{c}{Structure and Shape} 
    & \multicolumn{2}{c}{Surface Normal} 
    & \multicolumn{1}{c}{Edges} 
    \\

    \multicolumn{1}{c|}{} 
    & RMSE $\downarrow$ 
    & Abs. $\downarrow$ 
    & $\delta_1$ $\uparrow$
    & $\angle$~Dist $\downarrow$ 
    & \% wtn $t^{\circ}$ $\uparrow$
    & $\text{D}^{3}\text{R}$ $\downarrow$ 
    
    & RMSE $\downarrow$ 
    & Abs. $\downarrow$ 
    & $\delta_1$ $\uparrow$ 
    & $\angle$~Dist $\downarrow$ 
    & \% wtn $t^{\circ}$ $\uparrow$
    & $\text{D}^{3}\text{R}$ $\downarrow$ 
    & $\varepsilon_{\text{DBE}}^{\text{acc}}$ $\downarrow$ 
    & $\varepsilon_{\text{DBE}}^{\text{comp}}$ $\downarrow$ 
    
    & RMSE $\downarrow$ 
    & Abs. $\downarrow$ 
    & $\delta_1$ $\uparrow$  
    & $\angle$~Dist $\downarrow$ 
    & \% wtn $t^{\circ}$ $\uparrow$
    & $\text{D}^{3}\text{R}$ $\downarrow$
    \\
    \hline
    
    MD~\cite{li2018megadepth}
    & 58.4 & 39.4 & 44.2 & 74.9 & 16.6 & 0.552 
    & 2.20 & 47.2 & 48.6 & 51.1 & 12.1 & 0.596 & 3.29 & 58.4 
    & 2.92 & 57.2 & 42.3 & 53.3 & 7.80 & 0.857 \\ 
    
    MC~\cite{li2019learning}
    & 61.4 & 50.6 & 42.8 & 73.3 & 16.2 & 0.694 
    & 1.07 & 22.7 & 60.6 & 48.0 & 11.7 & 0.724 & 4.08 & 57.4 
    & 1.63 & 30.5 & 52.1 & 48.6 & 10.30 & 0.901 \\ 
    
    VN ICCV~\cite{yin2019enforcing}
    & 64.4 & 49.3 & 41.5 & 75.8 & 15.9 & 0.698 
    & \uline{0.74} & \uline{13.7} & \uline{80.4}  & 39.9 & 22.8 & 0.707 & 4.27 & 30.9 
    & \uline{0.97} & \uline{16.0} & \uline{77.2} & 39.1 & 21.2 & 0.910 \\ 
    
    LeReS~\cite{leres}
    & \uline{42.6} & \uline{34.3} & \topscore{56.0} & \uline{65.1} & \uline{22.7} & \uline{0.415} 
    & 0.88 & 20.2 & 68.7 & \topscore{25.3} & \topscore{42.1} & \uline{0.431} & \uline{2.25} & \uline{20.1} 
    & 1.49 & 27.3 & 56.1 & \uline{28.9} & \uline{32.0} & \uline{0.745} \\ 

    \hline
    Ours SI 
    & \topscore{41.3} & \topscore{34.0} & \uline{55.4} & \topscore{58.4} & \topscore{24.1} & \topscore{0.215} 
    & \topscore{0.69} & \topscore{11.7} & \topscore{86.7} & \uline{26.9} & \uline{35.1} & \topscore{0.342} & \topscore{1.69} & \topscore{16.0} 
    & \topscore{0.89} & \topscore{15.7} & \topscore{80.1} & \topscore{26.0} & \topscore{36.8} &  \topscore{0.742} \\ 

    \hline
    \end{tabular}
}
}
\end{table*}

%% file: tex/5_experiments.tex
\input{figures/comp_leres_inthewild}

We present an evaluation of our method using datasets that were not included in the training, namely Middlebury2014 \cite{scharstein2014high}, iBims1 \cite{koch2018evaluation}, and DIODE \cite{vasiljevic2019diode}-(indoor).

\subsection{SI depth evaluation}
To perform a comprehensive numerical evaluation of our SI depth, we have selected three categories of metrics.
(I) RMSE, Absolute relative difference (Abs.) and $\delta_1= \max(\frac{z}{z^*}, \frac{z^*}
{z}) < 1.25$ assess shape and structure of the scene.
(II) Angle Distance ($\angle~Dist$) \cite{chen2017surface} and $\%~wtn~t$ \cite{chen2017surface} focus on surface orientation accuracy.
 (III) $D^3R$ \cite{bmd}, $\varepsilon_{\text{DBE}}^{\text{comp}}$ and $\varepsilon_{\text{DBE}}^{\text{acc}}$ \cite{koch2018evaluation} measure depth discontinuities, edge completeness and location accuracy, respectively.

The results, presented in Table~\ref{tab:si_zeroshot}, demonstrate that our method significantly enhances the accuracy of scale-invariant depth estimation across various metrics.
Compared to the current state-of-the-art (SOTA) techniques, our method consistently produces superior structures, depth distribution, and boundary accuracy.
Moreover, the evaluation of surface orientation metrics indicates that our method outperforms existing approaches when applied to high-resolution datasets like Middlebury and DIODE, while achieving competitive performance on the iBims1 dataset.

We also present qualitative comparisons of our method to SOTA in Figure~\ref{fig:teaser}, ~\ref{fig:comparison:all:depth}, \ref{fig:comparison:leres:pcd3view}, and \ref{fig:comparison:leres:inthewild}. Our results exhibit significantly improved boundary localization compared to the competing networks. To visualize the reconstructed shape and structure of the scene, we project the images into 3D point clouds using our depth estimations and compare them to the results obtained by LeReS for a variety of in-the-wild images in Figure~\ref{fig:teaser}, \ref{fig:comparison:leres:pcd3view}, and \ref{fig:comparison:leres:inthewild}. We use the focal length values estimated by LeReS for projection.  Our results in Figure~\ref{fig:comparison:leres:pcd3view} are provided in various angles to demonstrate the precise scene shape and structure generated by our method in addition to the accurately captured object boundaries. 
In contrast, LeReS by only relying on a low-resolution SSI depth as input, fails to accurately detect many object boundaries, resulting in an inadequate representation of the complex scenes. Our method's high level of detail and accurate geometry enables the use of SI depth in 3D photography applications. We provide results for the 3D Photography task~\cite{shih20203d} with comparisons to other SI MDE methods as well as further qualitative examples in the supplementary material.

\input{tables/metric_zeroshot}

\subsubsection{Metric Depth Estimation Methods}

Metric depth estimation models aim to go beyond SI depth by removing the arbitrary scale and defining depth in meters. However, when the focal length is unknown or not accounted for in the metric setup, these models tend to generate results with a scale mismatch. We summarize our comparison against metric depth estimation models in Table~\ref{tab:metric_zeroshot}. PatchFusion~\cite{patchfusion} and Zoedepth~\cite{zoedepth} perform poorly when evaluated on unseen datasets of iBims1 and Middlebury2014 due to the mismatch between their training focal length and images from these datasets. Metric3D~\cite{metric3d}, on the other hand, takes the camera parameters into account and generates comparable results on iBims1 dataset. However, it fails to faithfully recover metric depth for high-resolution and complex dataset of Middlebury2014. To ensure a fair comparison, we match the scale outputs to that of the ground truth and include these scale invariant evaluations in Table~\ref{tab:metric_zeroshot} as well. 

ZoeDepth~\cite{zoedepth}, which also employs SSI depth as input to estimate metric depth, fails to match our performance in detail generation. We believe this is due to our SSI-network's superior detail generation and our adaptation of multi-resolution processing. Metric3D~\cite{metric3d} achieves the low edge metric by estimating SI-depth in a single forward pass, reflecting the task's complexity. PatchFusion shows better edge scores than ours, while performing worse in overall structure. As we will discuss in the supplementary material, PatchFusion uses a model with 700M parameters, compared to our 180M parameters, and takes 60 times longer to process at 3 minutes vs. 3 seconds average, due to its patch-based iterative approach. Our approach recovers high-quality details with only 3 forward passes across our pipeline in a faster runtime that is comparable to other state-of-the-art methods.

As Table~\ref{tab:metric_zeroshot} shows, metric MDE models show a significant drop in performance in the Middlebury dataset when compared to their results on iBims-1. 
With its high-resolution ground-truth in complex environments, zero-shot evaluation on the Middlebury dataset is more challenging for metric or SI formulations where the training data is limited. 
Our superior performance on this dataset demonstrates the effectiveness of the SSI input in our formulation, which allows us to generalize to complex scenes even with synthetic-only training of SI MDE.

\input{tables/ord_zeroshot}

\input{figures/comp_leres_3view}

\subsection{SSI depth evaluation}
We evaluate the performance of our SSI depth estimation module against SOTA methods, as detailed in Table~\ref{tab:ord_zeroshot}.
The assessment includes metrics such as $D^3R$ \cite{bmd}, $\varepsilon_{\text{DBE}}^{\text{comp}}$, and $\varepsilon_{\text{DBE}}^{\text{acc}}$ \cite{koch2018evaluation} to gauge the quality of depth discontinuities, which is the primary focus of our work.
Additionally, to measure the structural coherence of the estimated depth map, we employ the ordinal relation metric (ORD) proposed by \citet{sgr}.

Table~\ref{tab:ord_zeroshot} demonstrates that our method consistently outperforms other baselines in generating detailed depth maps, benefiting from our novel loss combination, except when compared to DepthAnything~\cite{depthanything}. However, employing a CNN backbone allows our method to be boosted by \citet{bmd}'s boosting framework which is not applicable to transformer-based DepthAnything and DPT~\cite{dpt}. Results indicate that our boosted method generates substantial amount of details, outperforming every baseline by a significant margin. The qualitative examples presented in Figure~\ref{fig:teaser} and \ref{fig:zero:ordinalComp} also illustrate the significant improvement of our SSI depth over other baselines in generating details.

However, this improvement in details comes with a slight degradation in the ORD metric representing overall structure. We believe this is the result of the limited network capacity which makes it harder for CNNs to maintain global coherency when generating details \cite{bmd}. 
When SSI inputs are utilized for SI depth, however, we see that our network with accurate details is more effective in providing important information for the SI network. We demonstrate in Section~\ref{sec:experiments:ablation:si} that the final SI-depth experiences a performance loss when our SSI model is substituted with another SOTA method. This underscores the high level of details fed to the SI depth model by our SSI method, making it more effective in simplifying the task of the SI network, leading to superior performance.

\subsection{Ablation studies}

\subsubsection{SSI MDE ablation}
\label{sec:experiments:ablation:ssi}
To assess the impact of our relaxed ranking loss on enabling mixed dense-sparse training, we conduct an ablation study.
Our study utilizes a subset of the Hypersim dataset \cite{roberts2021hypersim} consisting of 10,000 images (20 per scene) randomly sampled from the train split.
We train multiple networks for 20 epochs, employing different loss functions as outlined in Table~\ref{tab:ord_abl}.
Throughout all setups, we use $\mathcal{L}_{ssig}$ as a default component due to its crucial role in generating spatially coherent estimations.
The results support our discussion in Section~\ref{sec:method} that a naive combination of ranking and SSI loss yields inferior performance compared to utilizing either of them individually.
However, with our novel definition of the ordinal loss a higher performance is achieved when two losses are combined. 

\input{tables/si_abl}

\input{tables/ord_abl}

\subsubsection{SI MDE ablation}
\label{sec:experiments:ablation:si}
In order to assess the impact of using SSI depth estimations in the training of SI depth, we conduct an isolated test as summarized in Table~\ref{tab:si_abl}.
For this experiment, we train our scale-invariant network for 4 epochs with various settings and evaluated its performance.
Only using RGB as input demonstrates very low performance.
This shows a direct regression of SI-depth leads to a low performance due to the complexity of the task and limitations of neural networks. 

Feeding SSI depth estimations alongside RGB simplifies the SI network's task and enhances the SI depth model's capabilities, improving structure, surface normal, and detail metrics as reported in Table~\ref{tab:si_abl}.
Additionally, our model's SSI estimations outperform those from MiDaS, indicating their superiority in providing depth discontinuities and easing the scale-invariant depth estimation network's task. Despite MiDaS showing better ORD performance in Table~\ref{tab:ord_zeroshot}, this ablation underscores that a higher level of details does indeed simplify the task more effectively.

The results of the variant with omitted high-resolution depth estimations exhibit reduced performance in details, assessed by $D^3R$. Interestingly, it also indicates a decreased ability to estimate the structure and shape of the scene, evaluated by $Abs.$ and $\delta_1$. We believe as the network allocates its capacity to recover details, it compromises its ability to accurately recover the structure.

Finally, Table~\ref{tab:si_abl} indicates that the surface normal loss plays a crucial role in faithfully recovering the shape and structure of the scene. A surface normal loss promotes better shape and structure by penalizing incorrect surface orientations, especially on flat regions. The $D^3R$ evaluation shows that the network's ability to recover details is not heavily affected by the surface normal loss, as it only decreases marginally. This can be attributed to the dominance of flat surfaces in the surface normal loss, as they constitute the majority of the images.

%% file: figures/comp_leres_inthewild.tex
\begin{figure*}[p]
    \centering
    \includegraphics[width=\linewidth]{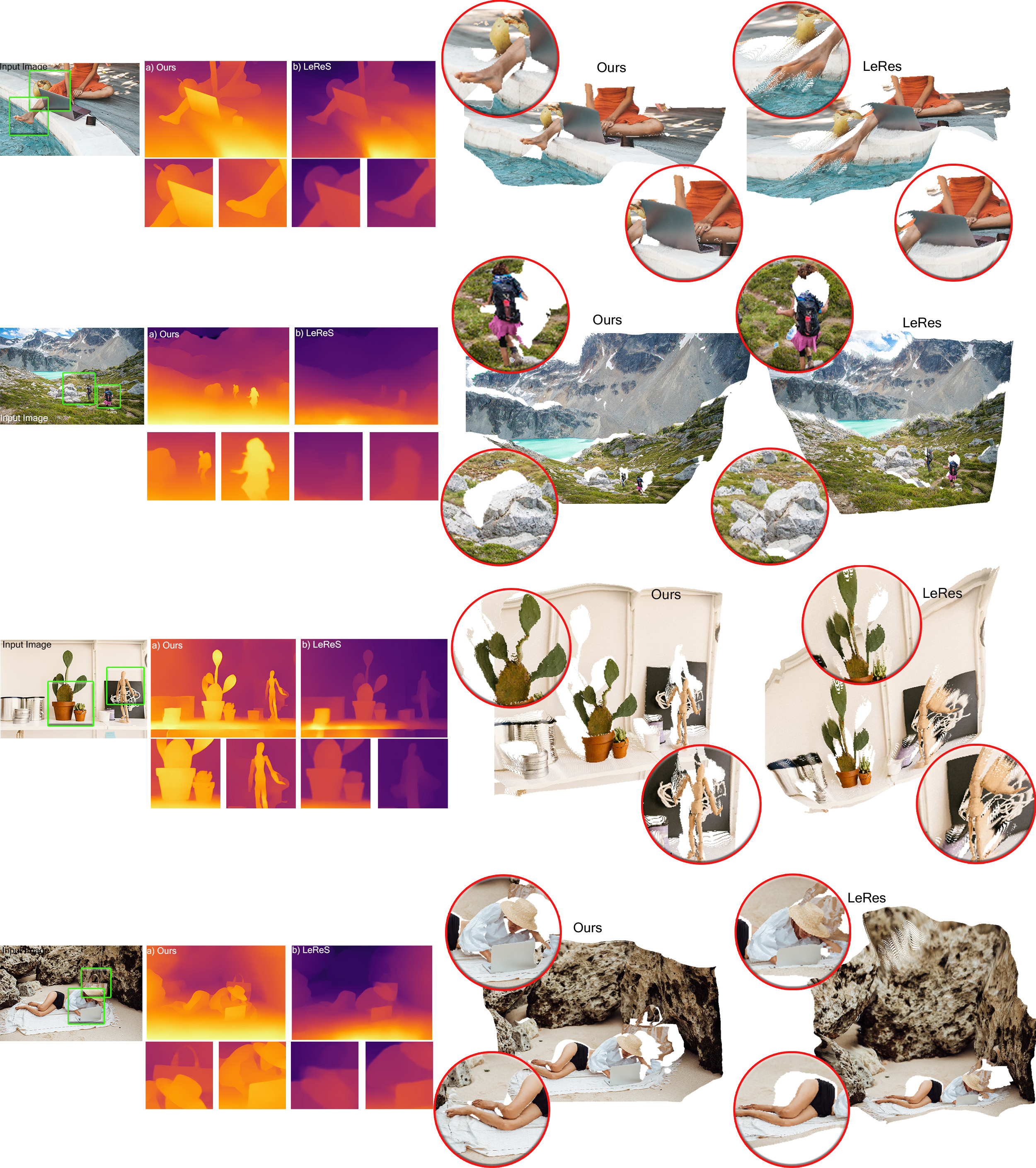}
    \caption{ Figure depicts the in-the-wild performance of our model in accurately modeling the scene compared to LeRes~\cite{leres}. Our model can model the 3D shape of various scenes with different depth distributions at a high resolution and with precise boundary accuracy. As highlighted by the insets, the absence of details in LeReS causes geometrical distortions in the projected point clouds. Our accurate boundary localization enables precise shape representation, even for complex in-the-wild scenes.
    \imagecredits{ 
    Death to the Stock Photo}
    }
\label{fig:comparison:leres:inthewild}
\end{figure*}

%% file: tables/metric_zeroshot.tex
\begin{table}[!t]
\caption{ Quantitative evaluation of metric depth estimation methods. These networks often inaccurately estimate depth due to focal length mismatch. Accurate results are achieved only after scale adjustment (marked with $\dagger$). }
\label{tab:metric_zeroshot}
\resizebox{\linewidth}{!}{%
{\renewcommand{\arraystretch}{1.2}
    \begin{tabular}{l|cccc|cccc}
    \hline
    Methods 
    & \multicolumn{4}{c|}{Middlebury} 
    & \multicolumn{4}{c}{iBims-1} 
    \\




    \multicolumn{1}{c|}{} 
    & RMSE $\downarrow$ 
    & Abs. $\downarrow$ 
    & $\delta_1$ $\uparrow$
    & $\text{D}^{3}\text{R}$ $\downarrow$ 
    
    & RMSE $\downarrow$ 
    & Abs. $\downarrow$ 
    & $\delta_1$ $\uparrow$ 
    & $\text{D}^{3}\text{R}$ $\downarrow$ 
    
    \\
    \hline

    Metric3D 
    & 218.6 & 186.8 & 58.9 & 0.443 
    & 0.60 & 17.5 & 79.5 & 19.3 
    \\

    Zoedepth  
    & 229.8 & 169.4 & 22.1 & 0.245 
    & 0.80  & 16.8 & 71.6  & 0.368 
    \\
    
    PatchFusion
    & 223.7 & 150.3 & 22.4 & 0.076 
    & 0.86  & 20.9  & 58.4 & 0.230  
    \\
    
    \hline

    Metric3D $\dagger$ 
    & 51.7 & 45.6 & 50.8 & 0.400 
    & \topscore{0.46} & 8.35 & \topscore{92.5} & 0.440 
    \\
    
    Zoedepth $\dagger$ 
    & 47.2 & 43.4 & \uline{56.8} & 0.239 
    & \uline{0.51} & \topscore{7.96} & \uline{92.4} & 0.369  
    \\

    PatchFusion $\dagger$ 
    & \uline{42.9} & \uline{40.4} & \topscore{58.1} & \topscore{0.076} 
    & 0.57 & 9.24 & 91.2 & \topscore{0.248}  
    \\

    \hline
    Ours SI
    & \topscore{41.3} & \topscore{34.0} & 55.4 & \uline{0.215} 
    & 0.69 & 11.7 & 86.7 & \uline{0.342}  
    \\ 

    \hline
    \end{tabular}
}
}
\vspace{-0.25cm}
\end{table}

%% file: tables/ord_zeroshot.tex
\begin{table}
\caption{Overview of the ordinal depth quantitative comparison. "bmd" indicates boosted using \cite{bmd}.}

\label{tab:ord_zeroshot}

\resizebox{\linewidth}{!}{%
{\renewcommand{\arraystretch}{1.2}
    \begin{tabular}{l|cc|ccc|cc}
    \hline
    Methods
    & \multicolumn{2}{c|}{Middlebury}
    & \multicolumn{3}{c}{iBims-1} 
    & \multicolumn{2}{c}{DIODE} 
    \\

    \multicolumn{1}{c|}{} 
    & Ord. $\downarrow$ 
    & $\text{D}^{3}\text{R}$ $\downarrow$  
    
    & Ord. $\downarrow$ 
    & $\text{D}^{3}\text{R}$ $\downarrow$ 
    & $\varepsilon_{\text{DBE}}^{\text{comp}}$
    ($\varepsilon_{\text{DBE}}^{\text{acc}})$    $\downarrow$  
    
    & Ord. $\downarrow$ 
    & $\text{D}^{3}\text{R}$ $\downarrow$ 

    \\ 
    \hline
    
     VN TPAMI~\cite{yin2021virtual} 
                & 0.213 & 0.613                
                & 0.140 & 0.623 & 52.9(4.68) 
                & 0.167 & 0.935 
                \\                           
    SGR~\cite{sgr}
                & 0.221 & 0.507                  
                & 0.200 & 0.522 & 34.6(2.37) 
                & 0.288 & 0.831 
                \\                                      
    Ken Burns~\cite{niklaus20193d} 
                & 0.221 & 0.453               
                & 0.125 & 0.487 & 22.8(2.19) 
                & 0.226 & 0.883 
                \\    
    
    LeReS SSI~\cite{leres}
                & 0.199 & 0.444               
                & 0.108  & 0.459 & 23.9(2.40) 
                & 0.143 & 0.820 
                \\                 
                
    MDS~\cite{midas} 
                & 0.176 & 0.449               
                & 0.128 & 0.458 & 28.4(2.22)  
                & 0.167 & 0.846 
                \\                     
                   
    DPT~\cite{dpt} 
                & \uline{0.162} & 0.369    
                & \uline{0.101} & 0.403 & 23.2(2.20) 
                & \uline{0.134} & 0.818 
                \\                    

    DepthAnything ~\cite{depthanything}
                & \topscore{0.092} & \uline{0.155}    
                & \topscore{0.051} & \uline{0.334} & \uline{12.5(1.94)} 
                & \topscore{0.074} & \topscore{0.771} 
                \\                    

    Our SSI
                & 0.190 & 0.339             
                & 0.112 & 0.345 & 22.5(2.16) 
                & 0.147 & 0.817 
                \\            
   \hline
   SGR-bmd
                & 0.210 & 0.280           
                & 0.196 & 0.411 & 22.3(2.29) 
                & 0.287 & 0.804 
                \\                                      
                
   MDS-bmd 
                & 0.162 & 0.201     
                & 0.126 & 0.368 & 23.3(2.15)  
                & 0.165 & 0.828 
                \\                                     
                
    \hline
    Ours bmd 
                & 0.174 & \topscore{0.120}             
                & 0.116 & \topscore{0.255} & \topscore{11.5(2.23)}  
                & 0.150 & \uline{0.790} 
                \\   
    \hline
    \end{tabular}
}
}
\vspace{-0.25cm}

\end{table}

%% file: figures/comp_leres_3view.tex
\begin{figure*}[p]
    \centering
    \includegraphics[width=\textwidth]{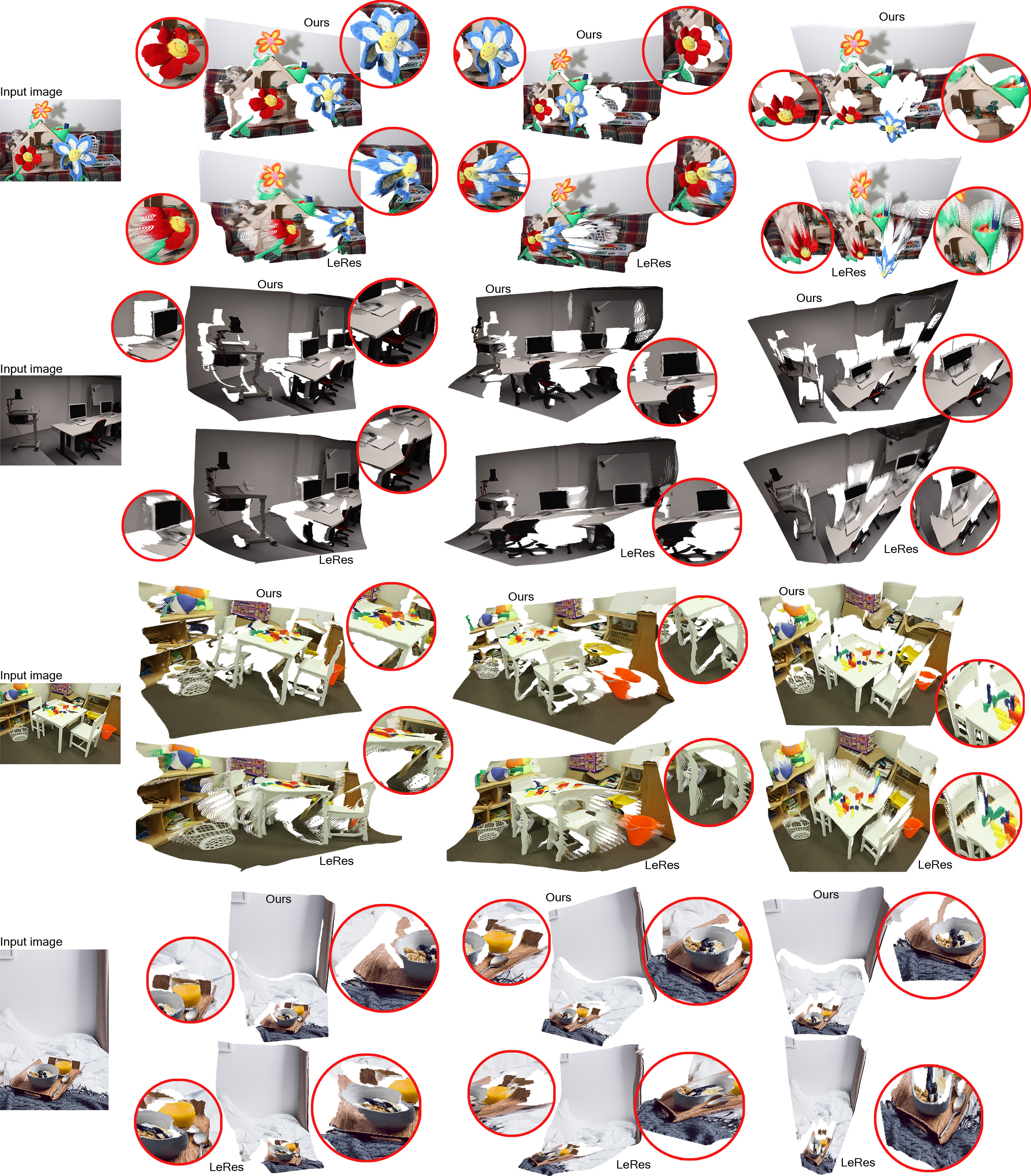}
    \caption{3D point clouds generated by our SI-depth and LeRes~\cite{leres} from various views shows leveraging our crisp SSI depth, our SI depth produces finer details. This results in a more precise representation of shape compared to the less detailed and inaccurate results of LeRes. The missing details in LeReS leads to distortion and blending of the details into the background. (see  flowers in the first row, monitors in the 2\ts{nd} row, objects on the table in the 3\ts{rd} row and the tray in the last row as emphasized by the insets.)
    \hfill
    \imagecredits{
     \cite{scharstein2014high}, \cite{koch2018evaluation}, Death to the Stock Photo
    }
    }
    \label{fig:comparison:leres:pcd3view}
\end{figure*}

%% file: tables/si_abl.tex
\begin{table}[!t]
    \caption{Overview of the influence of SSI depth in improving the performance of SI depth estimation.}
    \label{tab:si_abl}

 \resizebox{\linewidth}{!}{%
{
    \renewcommand{\arraystretch}{1.2}
    \begin{tabular}{l|cccc|cccc}
    \hline
    Methods 
    & \multicolumn{4}{c|}{Middlebury}
    & \multicolumn{4}{c}{iBims-1}
    \\

    & Abs. $\downarrow$ 
    & $\delta_1$ $\uparrow$ 
    & $\angle$~Dist $\downarrow$ 
    & $\text{D}^{3}\text{R}$ $\downarrow$ 
    
    & Abs. $\downarrow$ 
    & $\delta_1$ $\uparrow$ 
    & $\angle$~Dist $\downarrow$ 
    & $\text{D}^{3}\text{R}$ $\downarrow$ 


    \\ 
    \hline
   Only RGB
                & 53.5 & 38.3 & 74.9 & 0.728         
                & 26.4 & 65.4 & 46.9 & 0.715         
                \\ 
    ~ + MiDaS $\mathcal{O}^{L,H}$ 
                &  48.3 & 50.2 & 63.4 & 0.335        
                &  15.4 & 74.9 & 38.0 & 0.536        
                \\ 
    ~ + Our SSI $\mathcal{O}^{L,H}$ (Ours-si)
                & \topscore{36.2} &\topscore{58.8} & \topscore{61.1} & \topscore{0.285}         
                & \topscore{11.7} & \topscore{86.1} & \topscore{30.7} & \topscore{0.379}         
                \\ 
    \hline
   w/o ${O}^H$
                & 37.2 & 57.7 & \topscore{61.1} & 0.383         
                & 12.4 & 83.2 & 32.5 & 0.437         
                \\ 
    w/o Normal loss
                & 42.4 & 53.8 & 66.9 & 0.290         
                & 12.4 & 83.2 & 36.8 & 0.409         
                \\ 

    \end{tabular} 
}
}

\end{table}

%% file: tables/ord_abl.tex
\begin{table}[t]
\caption{Our ordinal loss can be combined with SSI yielding superior performance as opposed to ranking loss~\cite{chen2016single}.}
\vspace{-0.25cm}
\label{tab:ord_abl}
\center
\resizebox{0.7\linewidth}{!}{%
{\renewcommand{\arraystretch}{1.2}
    \begin{tabular}{l|cc|cc}
    \hline
    Methods 
    & \multicolumn{2}{c|}{Hypersim} 
    & \multicolumn{2}{c}{iBims-1} \\

    & Ord.$^*$ $\downarrow$ 
    & $\text{D}^{3}\text{R}$ $\downarrow$  
    
    & Ord. $\downarrow$ 
    & $\text{D}^{3}\text{R}$ $\downarrow$ 
    
    \\ \hline
    
    + $\mathcal{L}_{ssi}$
                & 0.185 & 0.496         
                & 0.156 & 0.520 \\ 
    + $\mathcal{L}_{ranking}$
                & 0.238 & 0.570         
                & 0.235 & 0.573 \\ 
    + $\mathcal{L}_{ssi} + \mathcal{L}_{ranking}$
                & 0.227 & 0.562         
                & 0.213 & 0.565 \\ 
    + $\mathcal{L}_{ssi} + \mathcal{L}_{so} ~ (ours)$
                & \topscore{0.184} & \topscore{0.486}    
                & \topscore{0.147} & \topscore{0.497} \\ 

    \hline
    \end{tabular}
}
}

\vspace{-0.3cm}

\end{table}

%% file: figures/comp_all_depth_SSI_ibims.tex
\begin{figure*}
    \centering
    \includegraphics[width=\linewidth]{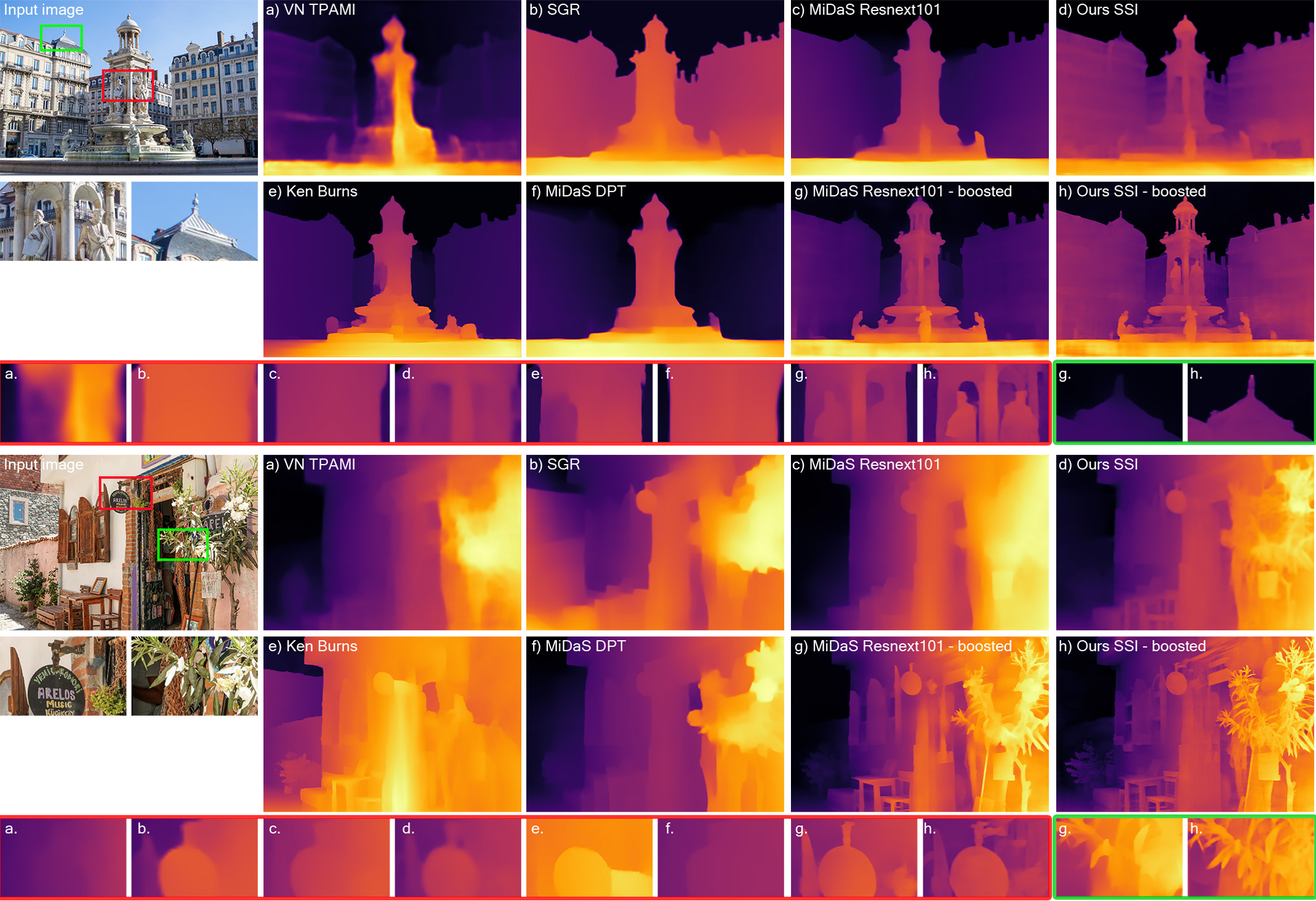}
    \caption{Qualitative comparison of scale and shift invariant networks in-the-wild reveals that our SSI network produces crisp depth boundaries compared to other methods. The results of our high-resolution boosted model exhibit even more refined depth boundaries.
    \imagecredits{
    \myhref[darkgray]{https://unsplash.com/photos/white-concrete-building-with-fountain-bNEaIT3HIMk}{@$\text{Diogo Nunes}$},
    \myhref[darkgray]{https://unsplash.com/photos/a-cafe-with-a-brick-building-Kl3yDaIY8nk}{@$\text{Mert Kahveci}$}   
    }
    }
    \label{fig:zero:ordinalComp}
\end{figure*}

%% file: tex/6_limitation.tex
Our method focuses on generating highly detailed, SI depth estimations. 
The quality of our estimations, however, depends on the quality of the input images. 
For low-resolutions, or noisy images, our method may fail to generate sharp results. 
This mainly comes from our high-resolution ordinal input failing to give accurate depth discontinuities in the case of image noise. We present further analysis and discussion on this in the supplementary material. 

We utilized a CNN architecture for both our SSI and SI depth estimation networks. 
This choice allows us to generate estimations at high resolutions with increasing details for our SSI network. However, the SI depth network struggles with increasing resolution due to global scale-invariant constraints, requiring the network to reason about every pixel in the image together. This constraint necessitates the entire input image to fit in our native resolution to generate consistent structures. This makes transformer-based architectures a good candidate for SI depth with SSI inputs. 
The lack of large SI depth datasets, however, creates a challenge for training a transformer-based architecture than for CNNs.

%% file: tex/7_conclusion.tex
We present a geometric monocular depth estimation method that can generate highly detailed and geometrically consistent reconstructions from a single image. 
To achieve this, we introduce an SSI depth estimation method that can generate sharper depth discontinuities. 
Using our SSI depth, we formulate SI MDE with SSI inputs, simplifying the SI MDE problem to the enforcement of geometric constraints. 
We show that through this simplification, in-the-wild generalization of SI task is achievable through only training with a synthetic indoors dataset, inheriting the generalization capability of SSI formulations that can be trained on diverse datasets,
Using our estimated SSI depth as input, we show that our novel scale-invariant depth estimation formulation can generate highly detailed results even for complex scenes in the wild. 
We have demonstrated state-of-the-art performance for scale-invariant depth estimation through zero-shot evaluations.

%% file: main.bbl

\begin{thebibliography}{52}


\ifx \showCODEN    \undefined \def \showCODEN     #1{\unskip}     \fi
\ifx \showDOI      \undefined \def \showDOI       #1{#1}\fi
\ifx \showISBNx    \undefined \def \showISBNx     #1{\unskip}     \fi
\ifx \showISBNxiii \undefined \def \showISBNxiii  #1{\unskip}     \fi
\ifx \showISSN     \undefined \def \showISSN      #1{\unskip}     \fi
\ifx \showLCCN     \undefined \def \showLCCN      #1{\unskip}     \fi
\ifx \shownote     \undefined \def \shownote      #1{#1}          \fi
\ifx \showarticletitle \undefined \def \showarticletitle #1{#1}   \fi
\ifx \showURL      \undefined \def \showURL       {\relax}        \fi
\providecommand\bibfield[2]{#2}
\providecommand\bibinfo[2]{#2}
\providecommand\natexlab[1]{#1}
\providecommand\showeprint[2][]{arXiv:#2}

\bibitem[Bhat et~al\mbox{.}(2021)]%
        {bhat2021adabins}
\bibfield{author}{\bibinfo{person}{Shariq~Farooq Bhat}, \bibinfo{person}{Ibraheem Alhashim}, {and} \bibinfo{person}{Peter Wonka}.} \bibinfo{year}{2021}\natexlab{}.
\newblock \showarticletitle{Adabins: Depth estimation using adaptive bins}. In \bibinfo{booktitle}{\emph{Proc. CVPR}}.
\newblock


\bibitem[Bhat et~al\mbox{.}(2022)]%
        {bhat2022localbin}
\bibfield{author}{\bibinfo{person}{Shariq~Farooq Bhat}, \bibinfo{person}{Ibraheem Alhashim}, {and} \bibinfo{person}{Peter Wonka}.} \bibinfo{year}{2022}\natexlab{}.
\newblock \showarticletitle{LocalBins: Improving Depth Estimation by Learning Local Distributions}. In \bibinfo{booktitle}{\emph{Proc. ECCV}}.
\newblock


\bibitem[Bhat et~al\mbox{.}(2023)]%
        {zoedepth}
\bibfield{author}{\bibinfo{person}{Shariq~Farooq Bhat}, \bibinfo{person}{Reiner Birkl}, \bibinfo{person}{Diana Wofk}, \bibinfo{person}{Peter Wonka}, {and} \bibinfo{person}{Matthias Müller}.} \bibinfo{year}{2023}\natexlab{}.
\newblock \showarticletitle{ZoeDepth: Zero-shot Transfer by Combining Relative and Metric Depth}.
\newblock \bibinfo{journal}{\emph{\tt arXiv:2302.12288 [cs.CV]}} (\bibinfo{year}{2023}).
\newblock


\bibitem[Chen et~al\mbox{.}(2016)]%
        {chen2016single}
\bibfield{author}{\bibinfo{person}{Weifeng Chen}, \bibinfo{person}{Zhao Fu}, \bibinfo{person}{Dawei Yang}, {and} \bibinfo{person}{Jia Deng}.} \bibinfo{year}{2016}\natexlab{}.
\newblock \showarticletitle{Single-image depth perception in the wild}. In \bibinfo{booktitle}{\emph{Proc. NeurIPS}}.
\newblock


\bibitem[Chen et~al\mbox{.}(2019b)]%
        {chen2019learning}
\bibfield{author}{\bibinfo{person}{Weifeng Chen}, \bibinfo{person}{Shengyi Qian}, {and} \bibinfo{person}{Jia Deng}.} \bibinfo{year}{2019}\natexlab{b}.
\newblock \showarticletitle{Learning single-image depth from videos using quality assessment networks}. In \bibinfo{booktitle}{\emph{Proc. CVPR}}.
\newblock


\bibitem[Chen et~al\mbox{.}(2017)]%
        {chen2017surface}
\bibfield{author}{\bibinfo{person}{Weifeng Chen}, \bibinfo{person}{Donglai Xiang}, {and} \bibinfo{person}{Jia Deng}.} \bibinfo{year}{2017}\natexlab{}.
\newblock \showarticletitle{Surface normals in the wild}. In \bibinfo{booktitle}{\emph{Proc. ICCV}}.
\newblock


\bibitem[Chen et~al\mbox{.}(2019a)]%
        {Chen2019structure-aware}
\bibfield{author}{\bibinfo{person}{Xiaotian Chen}, \bibinfo{person}{Xuejin Chen}, {and} \bibinfo{person}{Zheng-Jun Zha}.} \bibinfo{year}{2019}\natexlab{a}.
\newblock \showarticletitle{Structure-Aware Residual Pyramid Network for Monocular Depth Estimation}. In \bibinfo{booktitle}{\emph{Proc. IJCAI}}.
\newblock


\bibitem[Cheng et~al\mbox{.}(2022)]%
        {cheng2021masked}
\bibfield{author}{\bibinfo{person}{Bowen Cheng}, \bibinfo{person}{Ishan Misra}, \bibinfo{person}{Alexander~G Schwing}, \bibinfo{person}{Alexander Kirillov}, {and} \bibinfo{person}{Rohit Girdhar}.} \bibinfo{year}{2022}\natexlab{}.
\newblock \showarticletitle{Masked-attention mask transformer for universal image segmentation}. In \bibinfo{booktitle}{\emph{Proc. CVPR}}.
\newblock


\bibitem[Choi et~al\mbox{.}(2016)]%
        {choi2016large}
\bibfield{author}{\bibinfo{person}{Sungjoon Choi}, \bibinfo{person}{Qian-Yi Zhou}, \bibinfo{person}{Stephen Miller}, {and} \bibinfo{person}{Vladlen Koltun}.} \bibinfo{year}{2016}\natexlab{}.
\newblock \showarticletitle{A large dataset of object scans}.
\newblock \bibinfo{journal}{\emph{\tt arXiv:1602.02481 [cs.CV]}} (\bibinfo{year}{2016}).
\newblock


\bibitem[Eftekhar et~al\mbox{.}(2021)]%
        {eftekhar2021omnidata}
\bibfield{author}{\bibinfo{person}{Ainaz Eftekhar}, \bibinfo{person}{Alexander Sax}, \bibinfo{person}{Jitendra Malik}, {and} \bibinfo{person}{Amir Zamir}.} \bibinfo{year}{2021}\natexlab{}.
\newblock \showarticletitle{Omnidata: A Scalable Pipeline for Making Multi-Task Mid-Level Vision Datasets from 3D Scans}. In \bibinfo{booktitle}{\emph{Proc. ICCV}}.
\newblock


\bibitem[Eigen and Fergus(2015)]%
        {eigen2015predicting}
\bibfield{author}{\bibinfo{person}{David Eigen} {and} \bibinfo{person}{Rob Fergus}.} \bibinfo{year}{2015}\natexlab{}.
\newblock \showarticletitle{Predicting depth, surface normals and semantic labels with a common multi-scale convolutional architecture}. In \bibinfo{booktitle}{\emph{Proc. ICCV}}.
\newblock


\bibitem[Eigen et~al\mbox{.}(2014)]%
        {eigen2014depth}
\bibfield{author}{\bibinfo{person}{David Eigen}, \bibinfo{person}{Christian Puhrsch}, {and} \bibinfo{person}{Rob Fergus}.} \bibinfo{year}{2014}\natexlab{}.
\newblock \showarticletitle{Depth map prediction from a single image using a multi-scale deep network}. In \bibinfo{booktitle}{\emph{Proc. NeurIPS}}.
\newblock


\bibitem[Fu et~al\mbox{.}(2018)]%
        {dorn2018}
\bibfield{author}{\bibinfo{person}{Huan Fu}, \bibinfo{person}{Mingming Gong}, \bibinfo{person}{Chaohui Wang}, \bibinfo{person}{Kayhan Batmanghelich}, {and} \bibinfo{person}{Dacheng Tao}.} \bibinfo{year}{2018}\natexlab{}.
\newblock \showarticletitle{Deep ordinal regression network for monocular depth estimation}. In \bibinfo{booktitle}{\emph{Proc. CVPR}}.
\newblock


\bibitem[Godard et~al\mbox{.}(2017)]%
        {godard2017unsupervised}
\bibfield{author}{\bibinfo{person}{Cl{\'e}ment Godard}, \bibinfo{person}{Oisin Mac~Aodha}, {and} \bibinfo{person}{Gabriel~J Brostow}.} \bibinfo{year}{2017}\natexlab{}.
\newblock \showarticletitle{Unsupervised monocular depth estimation with left-right consistency}. In \bibinfo{booktitle}{\emph{Proc. CVPR}}.
\newblock


\bibitem[Hua et~al\mbox{.}(2020)]%
        {hua2020holopix50k}
\bibfield{author}{\bibinfo{person}{Yiwen Hua}, \bibinfo{person}{Puneet Kohli}, \bibinfo{person}{Pritish Uplavikar}, \bibinfo{person}{Anand Ravi}, \bibinfo{person}{Saravana Gunaseelan}, \bibinfo{person}{Jason Orozco}, {and} \bibinfo{person}{Edward Li}.} \bibinfo{year}{2020}\natexlab{}.
\newblock \showarticletitle{Holopix50k: A large-scale in-the-wild stereo image dataset}. In \bibinfo{booktitle}{\emph{Proc. CVPR Workshops}}.
\newblock


\bibitem[Jun et~al\mbox{.}(2022)]%
        {jun2022depth}
\bibfield{author}{\bibinfo{person}{Jinyoung Jun}, \bibinfo{person}{Jae-Han Lee}, \bibinfo{person}{Chul Lee}, {and} \bibinfo{person}{Chang-Su Kim}.} \bibinfo{year}{2022}\natexlab{}.
\newblock \showarticletitle{Depth map decomposition for monocular depth estimation}. In \bibinfo{booktitle}{\emph{Proc. ECCV}}.
\newblock


\bibitem[Koch et~al\mbox{.}(2018)]%
        {koch2018evaluation}
\bibfield{author}{\bibinfo{person}{Tobias Koch}, \bibinfo{person}{Lukas Liebel}, \bibinfo{person}{Friedrich Fraundorfer}, {and} \bibinfo{person}{Marco Korner}.} \bibinfo{year}{2018}\natexlab{}.
\newblock \showarticletitle{Evaluation of {CNN}-based single-image depth estimation methods}. In \bibinfo{booktitle}{\emph{Proc. ECCV Workshops}}.
\newblock


\bibitem[Kr{\"a}henb{\"u}hl(2018)]%
        {krahenbuhl2018free}
\bibfield{author}{\bibinfo{person}{Philipp Kr{\"a}henb{\"u}hl}.} \bibinfo{year}{2018}\natexlab{}.
\newblock \showarticletitle{Free supervision from video games}. In \bibinfo{booktitle}{\emph{Proc. CVPR}}.
\newblock


\bibitem[Lee and Kim(2019)]%
        {Jae2019depthusingrel}
\bibfield{author}{\bibinfo{person}{Jae-Han Lee} {and} \bibinfo{person}{Chang-Su Kim}.} \bibinfo{year}{2019}\natexlab{}.
\newblock \showarticletitle{Monocular Depth Estimation Using Relative Depth Maps}. In \bibinfo{booktitle}{\emph{Proc. CVPR}}.
\newblock


\bibitem[Li et~al\mbox{.}(2024)]%
        {patchfusion}
\bibfield{author}{\bibinfo{person}{Zhenyu Li}, \bibinfo{person}{Shariq~Farooq Bhat}, {and} \bibinfo{person}{Peter Wonka}.} \bibinfo{year}{2024}\natexlab{}.
\newblock \showarticletitle{PatchFusion: An End-to-End Tile-Based Framework for High-Resolution Monocular Metric Depth Estimation}. In \bibinfo{booktitle}{\emph{Proc. CVPR}}.
\newblock


\bibitem[Li et~al\mbox{.}(2019)]%
        {li2019learning}
\bibfield{author}{\bibinfo{person}{Zhengqi Li}, \bibinfo{person}{Tali Dekel}, \bibinfo{person}{Forrester Cole}, \bibinfo{person}{Richard Tucker}, \bibinfo{person}{Noah Snavely}, \bibinfo{person}{Ce Liu}, {and} \bibinfo{person}{William~T Freeman}.} \bibinfo{year}{2019}\natexlab{}.
\newblock \showarticletitle{Learning the depths of moving people by watching frozen people}. In \bibinfo{booktitle}{\emph{Proc. CVPR}}.
\newblock


\bibitem[Li and Snavely(2018)]%
        {li2018megadepth}
\bibfield{author}{\bibinfo{person}{Zhengqi Li} {and} \bibinfo{person}{Noah Snavely}.} \bibinfo{year}{2018}\natexlab{}.
\newblock \showarticletitle{Megadepth: Learning single-view depth prediction from internet photos}. In \bibinfo{booktitle}{\emph{Proc. CVPR}}.
\newblock


\bibitem[Li et~al\mbox{.}(2021)]%
        {li2021openrooms}
\bibfield{author}{\bibinfo{person}{Zhengqin Li}, \bibinfo{person}{Ting-Wei Yu}, \bibinfo{person}{Shen Sang}, \bibinfo{person}{Sarah Wang}, \bibinfo{person}{Meng Song}, \bibinfo{person}{Yuhan Liu}, \bibinfo{person}{Yu-Ying Yeh}, \bibinfo{person}{Rui Zhu}, \bibinfo{person}{Nitesh Gundavarapu}, \bibinfo{person}{Jia Shi}, {et~al\mbox{.}}} \bibinfo{year}{2021}\natexlab{}.
\newblock \showarticletitle{OpenRooms: An Open Framework for Photorealistic Indoor Scene Datasets}. In \bibinfo{booktitle}{\emph{Proc. CVPR}}.
\newblock


\bibitem[Mahajan et~al\mbox{.}(2018)]%
        {mahajan2018exploring}
\bibfield{author}{\bibinfo{person}{Dhruv Mahajan}, \bibinfo{person}{Ross Girshick}, \bibinfo{person}{Vignesh Ramanathan}, \bibinfo{person}{Kaiming He}, \bibinfo{person}{Manohar Paluri}, \bibinfo{person}{Yixuan Li}, \bibinfo{person}{Ashwin Bharambe}, {and} \bibinfo{person}{Laurens Van Der~Maaten}.} \bibinfo{year}{2018}\natexlab{}.
\newblock \showarticletitle{Exploring the limits of weakly supervised pretraining}. In \bibinfo{booktitle}{\emph{Proc. ECCV}}.
\newblock


\bibitem[Miangoleh et~al\mbox{.}(2021)]%
        {bmd}
\bibfield{author}{\bibinfo{person}{S~Mahdi~H Miangoleh}, \bibinfo{person}{Sebastian Dille}, \bibinfo{person}{Long Mai}, \bibinfo{person}{Sylvain Paris}, {and} \bibinfo{person}{Ya\u{g}{\i}z Aksoy}.} \bibinfo{year}{2021}\natexlab{}.
\newblock \showarticletitle{Boosting Monocular Depth Estimation Models to High-Resolution via Content-Adaptive Multi-Resolution Merging}. In \bibinfo{booktitle}{\emph{Proc. CVPR}}.
\newblock


\bibitem[Niklaus et~al\mbox{.}(2019)]%
        {niklaus20193d}
\bibfield{author}{\bibinfo{person}{Simon Niklaus}, \bibinfo{person}{Long Mai}, \bibinfo{person}{Jimei Yang}, {and} \bibinfo{person}{Feng Liu}.} \bibinfo{year}{2019}\natexlab{}.
\newblock \showarticletitle{3{D} {K}en {B}urns effect from a single image}.
\newblock \bibinfo{journal}{\emph{ACM Trans. Graph.}} (\bibinfo{year}{2019}).
\newblock


\bibitem[Peng et~al\mbox{.}(2022)]%
        {Peng2022BokehMe}
\bibfield{author}{\bibinfo{person}{Juewen Peng}, \bibinfo{person}{Zhiguo Cao}, \bibinfo{person}{Xianrui Luo}, \bibinfo{person}{Hao Lu}, \bibinfo{person}{Ke Xian}, {and} \bibinfo{person}{Jianming Zhang}.} \bibinfo{year}{2022}\natexlab{}.
\newblock \showarticletitle{BokehMe: When Neural Rendering Meets Classical Rendering}. In \bibinfo{booktitle}{\emph{Proc. CVPR}}.
\newblock


\bibitem[Ramamonjisoa et~al\mbox{.}(2020)]%
        {Ramamonjisoa_2020_CVPR}
\bibfield{author}{\bibinfo{person}{Michael Ramamonjisoa}, \bibinfo{person}{Yuming Du}, {and} \bibinfo{person}{Vincent Lepetit}.} \bibinfo{year}{2020}\natexlab{}.
\newblock \showarticletitle{Predicting Sharp and Accurate Occlusion Boundaries in Monocular Depth Estimation Using Displacement Fields}. In \bibinfo{booktitle}{\emph{Proc. CVPR}}.
\newblock


\bibitem[Ranftl et~al\mbox{.}(2021)]%
        {dpt}
\bibfield{author}{\bibinfo{person}{Ren{\'e} Ranftl}, \bibinfo{person}{Alexey Bochkovskiy}, {and} \bibinfo{person}{Vladlen Koltun}.} \bibinfo{year}{2021}\natexlab{}.
\newblock \showarticletitle{Vision transformers for dense prediction}. In \bibinfo{booktitle}{\emph{Proc. ICCV}}.
\newblock


\bibitem[Ranftl et~al\mbox{.}(2020)]%
        {midas}
\bibfield{author}{\bibinfo{person}{Ren{\'e} Ranftl}, \bibinfo{person}{Katrin Lasinger}, \bibinfo{person}{David Hafner}, \bibinfo{person}{Konrad Schindler}, {and} \bibinfo{person}{Vladlen Koltun}.} \bibinfo{year}{2020}\natexlab{}.
\newblock \showarticletitle{Towards Robust Monocular Depth Estimation: Mixing Datasets for Zero-shot Cross-dataset Transfer}.
\newblock \bibinfo{journal}{\emph{IEEE Trans. Pattern Anal. Mach. Intell.}} (\bibinfo{year}{2020}).
\newblock


\bibitem[Roberts et~al\mbox{.}(2021)]%
        {roberts2021hypersim}
\bibfield{author}{\bibinfo{person}{Mike Roberts}, \bibinfo{person}{Jason Ramapuram}, \bibinfo{person}{Anurag Ranjan}, \bibinfo{person}{Atulit Kumar}, \bibinfo{person}{Miguel~Angel Bautista}, \bibinfo{person}{Nathan Paczan}, \bibinfo{person}{Russ Webb}, {and} \bibinfo{person}{Joshua~M Susskind}.} \bibinfo{year}{2021}\natexlab{}.
\newblock \showarticletitle{Hypersim: A photorealistic synthetic dataset for holistic indoor scene understanding}. In \bibinfo{booktitle}{\emph{Proc. ICCV}}.
\newblock


\bibitem[Scharstein et~al\mbox{.}(2014)]%
        {scharstein2014high}
\bibfield{author}{\bibinfo{person}{Daniel Scharstein}, \bibinfo{person}{Heiko Hirschm{\"u}ller}, \bibinfo{person}{York Kitajima}, \bibinfo{person}{Greg Krathwohl}, \bibinfo{person}{Nera Ne{\v{s}}i{\'c}}, \bibinfo{person}{Xi Wang}, {and} \bibinfo{person}{Porter Westling}.} \bibinfo{year}{2014}\natexlab{}.
\newblock \showarticletitle{High-resolution stereo datasets with subpixel-accurate ground truth}. In \bibinfo{booktitle}{\emph{Proc. GCPR}}.
\newblock


\bibitem[Shih et~al\mbox{.}(2020)]%
        {shih20203d}
\bibfield{author}{\bibinfo{person}{Meng-Li Shih}, \bibinfo{person}{Shih-Yang Su}, \bibinfo{person}{Johannes Kopf}, {and} \bibinfo{person}{Jia-Bin Huang}.} \bibinfo{year}{2020}\natexlab{}.
\newblock \showarticletitle{3{D} photography using context-aware layered depth inpainting}. In \bibinfo{booktitle}{\emph{Proc. CVPR}}.
\newblock


\bibitem[Straub et~al\mbox{.}(2019)]%
        {straub2019replica}
\bibfield{author}{\bibinfo{person}{Julian Straub}, \bibinfo{person}{Thomas Whelan}, \bibinfo{person}{Lingni Ma}, \bibinfo{person}{Yufan Chen}, \bibinfo{person}{Erik Wijmans}, \bibinfo{person}{Simon Green}, \bibinfo{person}{Jakob~J Engel}, \bibinfo{person}{Raul Mur-Artal}, \bibinfo{person}{Carl Ren}, \bibinfo{person}{Shobhit Verma}, {et~al\mbox{.}}} \bibinfo{year}{2019}\natexlab{}.
\newblock \showarticletitle{The Replica dataset: A digital replica of indoor spaces}.
\newblock \bibinfo{journal}{\emph{\tt arXiv:1906.05797 [cs.CV]}} (\bibinfo{year}{2019}).
\newblock


\bibitem[Tan and Le(2019)]%
        {tan2019efficientnet}
\bibfield{author}{\bibinfo{person}{Mingxing Tan} {and} \bibinfo{person}{Quoc Le}.} \bibinfo{year}{2019}\natexlab{}.
\newblock \showarticletitle{Efficientnet: Rethinking model scaling for convolutional neural networks}. In \bibinfo{booktitle}{\emph{Proc. ICML}}.
\newblock


\bibitem[Teed and Deng(2020)]%
        {teed2020raft}
\bibfield{author}{\bibinfo{person}{Zachary Teed} {and} \bibinfo{person}{Jia Deng}.} \bibinfo{year}{2020}\natexlab{}.
\newblock \showarticletitle{Raft: Recurrent all-pairs field transforms for optical flow}. In \bibinfo{booktitle}{\emph{Proc. ECCV}}.
\newblock


\bibitem[Vasiljevic et~al\mbox{.}(2019)]%
        {vasiljevic2019diode}
\bibfield{author}{\bibinfo{person}{Igor Vasiljevic}, \bibinfo{person}{Nick Kolkin}, \bibinfo{person}{Shanyi Zhang}, \bibinfo{person}{Ruotian Luo}, \bibinfo{person}{Haochen Wang}, \bibinfo{person}{Falcon~Z Dai}, \bibinfo{person}{Andrea~F Daniele}, \bibinfo{person}{Mohammadreza Mostajabi}, \bibinfo{person}{Steven Basart}, \bibinfo{person}{Matthew~R Walter}, {et~al\mbox{.}}} \bibinfo{year}{2019}\natexlab{}.
\newblock \showarticletitle{Diode: A dense indoor and outdoor depth dataset}.
\newblock \bibinfo{journal}{\emph{\tt arXiv:1908.00463 [cs.CV]}} (\bibinfo{year}{2019}).
\newblock


\bibitem[Wadhwa et~al\mbox{.}(2018)]%
        {wadhwa2018synthetic}
\bibfield{author}{\bibinfo{person}{Neal Wadhwa}, \bibinfo{person}{Rahul Garg}, \bibinfo{person}{David~E Jacobs}, \bibinfo{person}{Bryan~E Feldman}, \bibinfo{person}{Nori Kanazawa}, \bibinfo{person}{Robert Carroll}, \bibinfo{person}{Yair Movshovitz-Attias}, \bibinfo{person}{Jonathan~T Barron}, \bibinfo{person}{Yael Pritch}, {and} \bibinfo{person}{Marc Levoy}.} \bibinfo{year}{2018}\natexlab{}.
\newblock \showarticletitle{Synthetic depth-of-field with a single-camera mobile phone}.
\newblock \bibinfo{journal}{\emph{ACM Trans. Graph.}} (\bibinfo{year}{2018}).
\newblock


\bibitem[Wang et~al\mbox{.}(2020a)]%
        {wang2020cliffnet}
\bibfield{author}{\bibinfo{person}{Lijun Wang}, \bibinfo{person}{Jianming Zhang}, \bibinfo{person}{Yifan Wang}, \bibinfo{person}{Huchuan Lu}, {and} \bibinfo{person}{Xiang Ruan}.} \bibinfo{year}{2020}\natexlab{a}.
\newblock \showarticletitle{{CLIFFNet} for Monocular Depth Estimation with Hierarchical Embedding Loss}. In \bibinfo{booktitle}{\emph{Proc. ECCV}}.
\newblock


\bibitem[Wang et~al\mbox{.}(2020b)]%
        {wang2020tartanair}
\bibfield{author}{\bibinfo{person}{Wenshan Wang}, \bibinfo{person}{Delong Zhu}, \bibinfo{person}{Xiangwei Wang}, \bibinfo{person}{Yaoyu Hu}, \bibinfo{person}{Yuheng Qiu}, \bibinfo{person}{Chen Wang}, \bibinfo{person}{Yafei Hu}, \bibinfo{person}{Ashish Kapoor}, {and} \bibinfo{person}{Sebastian Scherer}.} \bibinfo{year}{2020}\natexlab{b}.
\newblock \showarticletitle{Tartanair: A dataset to push the limits of visual {SLAM}}. In \bibinfo{booktitle}{\emph{Proc. IROS}}.
\newblock


\bibitem[Wong and Soatto(2019)]%
        {wong2019bilateral}
\bibfield{author}{\bibinfo{person}{Alex Wong} {and} \bibinfo{person}{Stefano Soatto}.} \bibinfo{year}{2019}\natexlab{}.
\newblock \showarticletitle{Bilateral cyclic constraint and adaptive regularization for unsupervised monocular depth prediction}. In \bibinfo{booktitle}{\emph{Proc. CVPR}}.
\newblock


\bibitem[Xian et~al\mbox{.}(2018)]%
        {xian2018monocular}
\bibfield{author}{\bibinfo{person}{Ke Xian}, \bibinfo{person}{Chunhua Shen}, \bibinfo{person}{Zhiguo Cao}, \bibinfo{person}{Hao Lu}, \bibinfo{person}{Yang Xiao}, \bibinfo{person}{Ruibo Li}, {and} \bibinfo{person}{Zhenbo Luo}.} \bibinfo{year}{2018}\natexlab{}.
\newblock \showarticletitle{Monocular relative depth perception with web stereo data supervision}. In \bibinfo{booktitle}{\emph{Proc. CVPR}}.
\newblock


\bibitem[Xian et~al\mbox{.}(2020)]%
        {sgr}
\bibfield{author}{\bibinfo{person}{Ke Xian}, \bibinfo{person}{Jianming Zhang}, \bibinfo{person}{Oliver Wang}, \bibinfo{person}{Long Mai}, \bibinfo{person}{Zhe Lin}, {and} \bibinfo{person}{Zhiguo Cao}.} \bibinfo{year}{2020}\natexlab{}.
\newblock \showarticletitle{Structure-guided ranking loss for single image depth prediction}. In \bibinfo{booktitle}{\emph{Proc. CVPR}}.
\newblock


\bibitem[Xie et~al\mbox{.}(2017)]%
        {xie2017aggregated}
\bibfield{author}{\bibinfo{person}{Saining Xie}, \bibinfo{person}{Ross Girshick}, \bibinfo{person}{Piotr Doll{\'a}r}, \bibinfo{person}{Zhuowen Tu}, {and} \bibinfo{person}{Kaiming He}.} \bibinfo{year}{2017}\natexlab{}.
\newblock \showarticletitle{Aggregated residual transformations for deep neural networks}. In \bibinfo{booktitle}{\emph{Proc. CVPR}}.
\newblock


\bibitem[Yang et~al\mbox{.}(2024)]%
        {depthanything}
\bibfield{author}{\bibinfo{person}{Lihe Yang}, \bibinfo{person}{Bingyi Kang}, \bibinfo{person}{Zilong Huang}, \bibinfo{person}{Xiaogang Xu}, \bibinfo{person}{Jiashi Feng}, {and} \bibinfo{person}{Hengshuang Zhao}.} \bibinfo{year}{2024}\natexlab{}.
\newblock \showarticletitle{Depth Anything: Unleashing the Power of Large-Scale Unlabeled Data}. In \bibinfo{booktitle}{\emph{Proc. CVPR}}.
\newblock


\bibitem[Yin et~al\mbox{.}(2021a)]%
        {yin2021virtual}
\bibfield{author}{\bibinfo{person}{Wei Yin}, \bibinfo{person}{Yifan Liu}, {and} \bibinfo{person}{Chunhua Shen}.} \bibinfo{year}{2021}\natexlab{a}.
\newblock \showarticletitle{Virtual Normal: Enforcing Geometric Constraints for Accurate and Robust Depth Prediction}.
\newblock \bibinfo{journal}{\emph{IEEE Trans. Pattern Anal. Mach. Intell.}} (\bibinfo{year}{2021}).
\newblock


\bibitem[Yin et~al\mbox{.}(2019)]%
        {yin2019enforcing}
\bibfield{author}{\bibinfo{person}{Wei Yin}, \bibinfo{person}{Yifan Liu}, \bibinfo{person}{Chunhua Shen}, {and} \bibinfo{person}{Youliang Yan}.} \bibinfo{year}{2019}\natexlab{}.
\newblock \showarticletitle{Enforcing geometric constraints of virtual normal for depth prediction}. In \bibinfo{booktitle}{\emph{Proc. ICCV}}.
\newblock


\bibitem[Yin et~al\mbox{.}(2023)]%
        {metric3d}
\bibfield{author}{\bibinfo{person}{Wei Yin}, \bibinfo{person}{Chi Zhang}, \bibinfo{person}{Hao Chen}, \bibinfo{person}{Zhipeng Cai}, \bibinfo{person}{Gang Yu}, \bibinfo{person}{Kaixuan Wang}, \bibinfo{person}{Xiaozhi Chen}, {and} \bibinfo{person}{Chunhua Shen}.} \bibinfo{year}{2023}\natexlab{}.
\newblock \showarticletitle{Metric3D: Towards Zero-shot Metric 3D Prediction from A Single Image}. In \bibinfo{booktitle}{\emph{Proc. ICCV}}.
\newblock


\bibitem[Yin et~al\mbox{.}(2021b)]%
        {leres}
\bibfield{author}{\bibinfo{person}{Wei Yin}, \bibinfo{person}{Jianming Zhang}, \bibinfo{person}{Oliver Wang}, \bibinfo{person}{Simon Niklaus}, \bibinfo{person}{Long Mai}, \bibinfo{person}{Simon Chen}, {and} \bibinfo{person}{Chunhua Shen}.} \bibinfo{year}{2021}\natexlab{b}.
\newblock \showarticletitle{Learning to recover 3d scene shape from a single image}. In \bibinfo{booktitle}{\emph{Proc. CVPR}}.
\newblock


\bibitem[Yuan et~al\mbox{.}(2022)]%
        {yuan2022newcrfs}
\bibfield{author}{\bibinfo{person}{Weihao Yuan}, \bibinfo{person}{Xiaodong Gu}, \bibinfo{person}{Zuozhuo Dai}, \bibinfo{person}{Siyu Zhu}, {and} \bibinfo{person}{Ping Tan}.} \bibinfo{year}{2022}\natexlab{}.
\newblock \showarticletitle{NeWCRFs: Neural Window Fully-connected CRFs for Monocular Depth Estimation}. In \bibinfo{booktitle}{\emph{Proc. CVPR}}.
\newblock


\bibitem[Zheng et~al\mbox{.}(2018)]%
        {zheng2018t2net}
\bibfield{author}{\bibinfo{person}{Chuanxia Zheng}, \bibinfo{person}{Tat-Jen Cham}, {and} \bibinfo{person}{Jianfei Cai}.} \bibinfo{year}{2018}\natexlab{}.
\newblock \showarticletitle{T2net: Synthetic-to-realistic translation for solving single-image depth estimation tasks}. In \bibinfo{booktitle}{\emph{Proc. ECCV}}.
\newblock


\bibitem[Zoran et~al\mbox{.}(2015)]%
        {midlevel}
\bibfield{author}{\bibinfo{person}{Daniel Zoran}, \bibinfo{person}{Phillip Isola}, \bibinfo{person}{Dilip Krishnan}, {and} \bibinfo{person}{William~T Freeman}.} \bibinfo{year}{2015}\natexlab{}.
\newblock \showarticletitle{Learning ordinal relationships for mid-level vision}. In \bibinfo{booktitle}{\emph{Proc. ICCV}}.
\newblock


\end{thebibliography}
